\documentclass[journal]{IEEEtran}%

\ifCLASSOPTIONcompsoc
\else
\fi
\ifCLASSINFOpdf
\else
\fi

\usepackage{times}
\usepackage{epsfig}
\usepackage{graphicx}
\usepackage{amsmath}
\usepackage{amssymb}
\usepackage{algorithm}
\usepackage{rotating}
\usepackage{algpseudocode}
\usepackage{booktabs}
\usepackage{url}

\DeclareMathOperator*{\argmax}{arg\,max}

\usepackage{tikz}
\usepackage{xparse}
\newsavebox\MyPicture
\NewDocumentCommand{\roundedpicture}%
      {O{width=0.6\linewidth}
       O{draw=blue,line width=.8pt,rounded corners=0pt}
       m}{%
   \savebox\MyPicture{\includegraphics[#1]{#3}}%
   \begin{tikzpicture}%
    \draw [path picture={%
                   \node at (path picture bounding box.center) {%
                       \usebox\MyPicture};},#2]
          (0,0)  rectangle (\wd\MyPicture,\ht\MyPicture);
   \end{tikzpicture}%
}

\begin{document}
%
\title{Scene Labeling with Contextual Hierarchical Models}
%
%
%
%

\author{Mojtaba~Seyedhosseini and
        Tolga~Tasdizen
\IEEEcompsocitemizethanks{\IEEEcompsocthanksitem M. Seyedhosseini and T. Tasdizen are with the Electrical and Computer Engineering Department and the Scientific Computing and Imaging Institute (SCI), University of Utah, Salt Lake City, UT, 84112 USA. \protect\\
E-mail: \{mseyed,tolga\}@sci.utah.edu 
}
\thanks{}}

\IEEEcompsoctitleabstractindextext{%
\begin{abstract}
Scene labeling is the problem of assigning an object label to each pixel.
It unifies the image segmentation and object recognition problems.
The importance of using contextual information in scene labeling frameworks has been widely realized in the field.
We propose a contextual framework, called \emph{contextual hierarchical model (CHM)}, which learns contextual information in a hierarchical framework for scene labeling.
At each level of the hierarchy, a classifier is trained based on downsampled input images and outputs of previous levels.
Our model then incorporates the resulting multi-resolution contextual information into a classifier to segment the input image at original resolution.
This training strategy allows for optimization of a joint posterior probability at multiple resolutions through the hierarchy.
Contextual hierarchical model is purely based on the input image patches and does not make use of any fragments or shape examples. Hence, it is applicable to a variety of problems such as object segmentation and edge detection.
We demonstrate that CHM outperforms state-of-the-art on Stanford background and Weizmann horse datasets. It also outperforms state-of-the-art edge detection methods on NYU depth dataset and achieves state-of-the-art on Berkeley segmentation dataset (BSDS 500).
\end{abstract}

\begin{keywords}
Scene Labeling, Image Segmentation, Edge Detection, Hierarchical Models, Membrane Detection, Connectome
\end{keywords}}

\maketitle

\IEEEdisplaynotcompsoctitleabstractindextext

%
\IEEEpeerreviewmaketitle

\section{Introduction}
%
%

%
%
%
%

\IEEEPARstart{S}{cene} labeling is of substantial importance for a wide range of applications in computer vision~\cite{Li1995}.
It is the primary step towards image understanding and integrates detection and segmentation in a single framework~\cite{Farabet2013}.
For instance, in a dataset of horse images, scene labeling can be thought of the task of labeling each pixel as part of a horse or non-horse, \textit{i.e.,} background.
In more complicated cases such as outdoor scene images, it might require multiple labels, \textit{e.g.,} buildings, cars, roads, sky etc.
This general definition can also be extended to the edge detection problem where each pixel is classified as edge or non-edge in a binary-decision framework.

Pixels can not be labeled based only on a small region around them.
For example, it is almost impossible to distinguish a pixel belonging to sky from a pixel belonging to sea by only looking at a small patch around them.
Therefore, a scene labeling framework needs to take into account short-range and long-range contextual information.
Contextual information has been widely used for solving high-level vision problems in computer vision~\cite{Tu2010,Torralba2004,Heitz2008,Li2012}.
Contextual information can refer to either inter-object configuration, \textit{e.g.,} a segmented horse's body may suggest the position of its legs~\cite{Tu2010}, or intra-object dependencies, \textit{e.g.,} the existence of a keyboard in an image implies that there is very likely a mouse near it~\cite{Torralba2004}.
From the Bayesian point of view, contextual information can be interpreted as the probability image map of an object, which carries prior information in the maximum aposteriori (MAP) pixel classification problem.

An important question about any scene labeling method is how it takes contextual information into account.
The main challenge is to pool contextual information from a large neighborhood while keeping the complexity tractable~\cite{Farabet2013}.
A common approach is to use a series of cascaded classifiers~\cite{Heitz2008,Tu2010,Li2012,Jurrus2010}.
In this architecture, each classifier is sequentially trained using the outputs of the previous classifiers as inputs.
This gradually increases the area of influence and allows later classifiers in the series to obtain contextual information from larger neighborhood areas.
However, they have a drawback that they do not obtain contextual information at multiple scales.
Multi-scale processing of images has been proven critical in many computer vision tasks~\cite{RenImage2013,SeyedhosseiniMulti2013}.
OWT-UCM~\cite{Arbelaez2009} takes advantage of processing the input image at multiple scales through a hierarchy.
This leads to state-of-the-art performance for edge detection applications.
Farabet \textit{et al.}~\cite{Farabet2013} showed that using multi-scale convolutional networks (ConvNets) can improve the performance of ConvNets dramatically for scene labeling.

This paper presents a contextual hierarchical model (CHM), which is able to obtain contextual information at multiple resolutions.
Similar to cascaded classifier models, CHM learns a series of classifiers consecutively, but unlike those models, it trains classifiers at multiple resolutions in a hierarchy.
The main advantage of CHM is that it targets a posterior probability at multiple resolutions and maximizes it greedily through the hierarchy.
This allows CHM to cover a large contextual window without adding intractable complexity.
While common approaches to scene labeling usually need postprocessing to ensure the consistency of labels, the use of a large contextual window reduces the requirement for sophisticated postprocessing methods.

A striking characteristic of our proposed method is that it is purely based on input image patches and does not make use of any shape fragments or object models, therefore, it is applicable to a wide range of applications such as edge detection and image labeling.
While some approaches such as~\cite{Arbelaez2009,Dollar2013,Ren2012,Catanzaro2009} can only be applied to edge detection problems and other approaches such as~\cite{Bertelli2011,Kuettel2012,RenRGB2012} are only designed for the image labeling problem, CHM can handle both problems equally well without any modification.

\begin{figure}
\def \lll{-.18cm}
\def \ll{-.37cm}
\begin{tabular}{ccc}
\hspace{\lll}\roundedpicture[width = .16\textwidth]{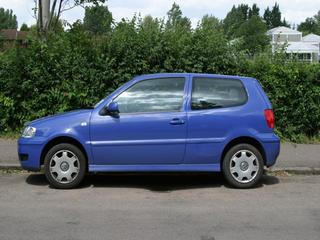}&
\hspace{\ll}\roundedpicture[width = .16\textwidth]{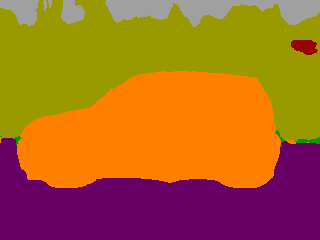}&
\hspace{\ll}\roundedpicture[width = .16\textwidth]{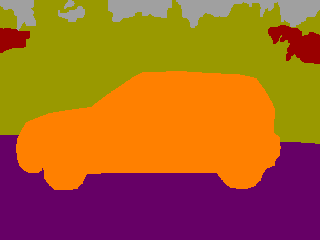}\\[-.5ex]
\hspace{\lll}\roundedpicture[width = .16\textwidth]{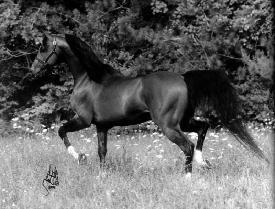}&
\hspace{\ll}\roundedpicture[width = .16\textwidth]{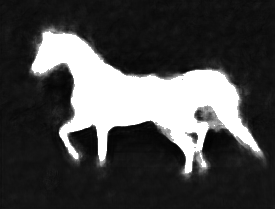}&
\hspace{\ll}\roundedpicture[width = .16\textwidth]{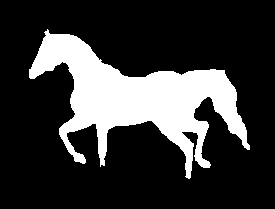}\\[-.5ex]
\hspace{\lll}\roundedpicture[width = .16\textwidth]{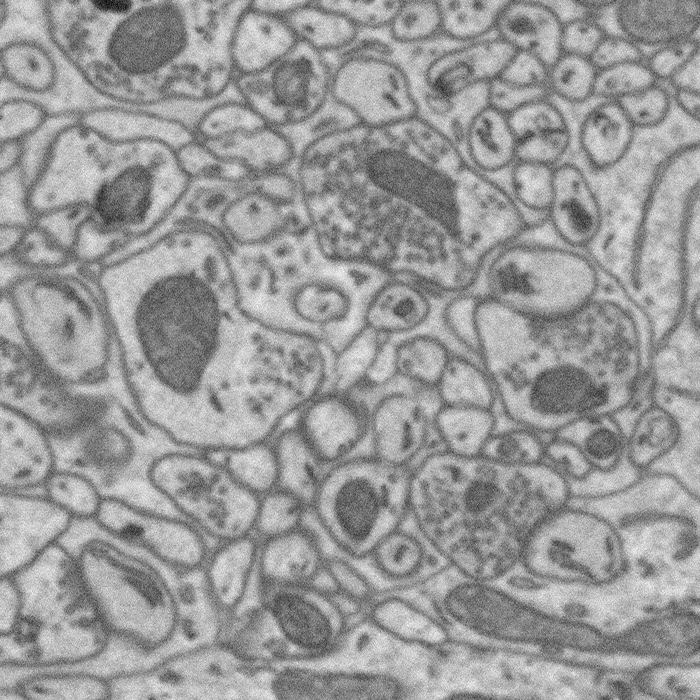}&
\hspace{\ll}\roundedpicture[width = .16\textwidth]{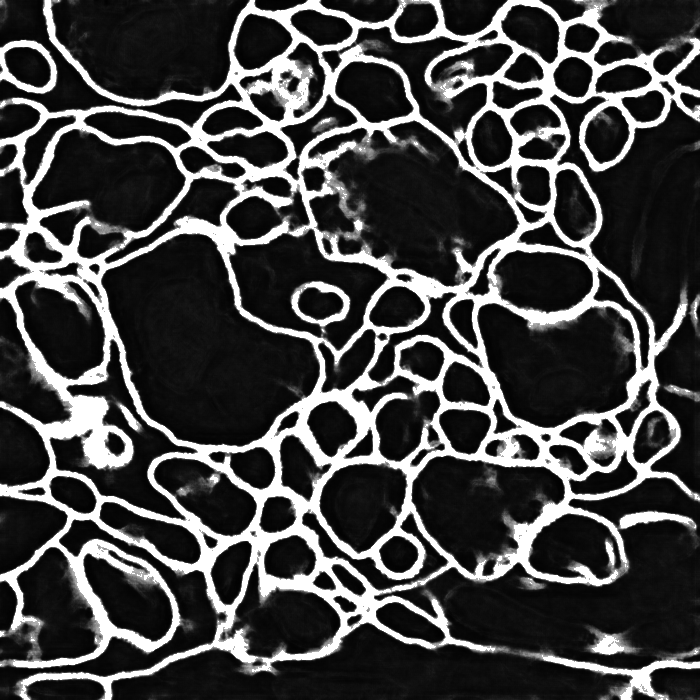}&
\hspace{\ll}\roundedpicture[width = .16\textwidth]{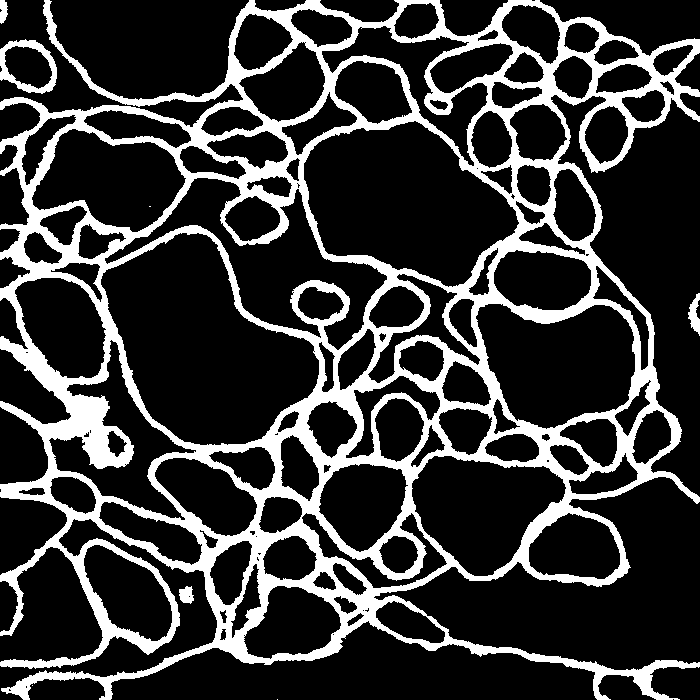}\\[-.5ex]
\hspace{\lll}\roundedpicture[width = .16\textwidth]{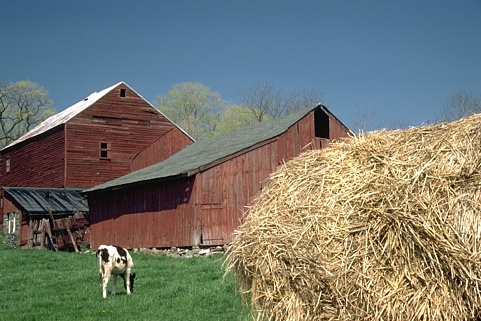}&
\hspace{\ll}\roundedpicture[width = .16\textwidth]{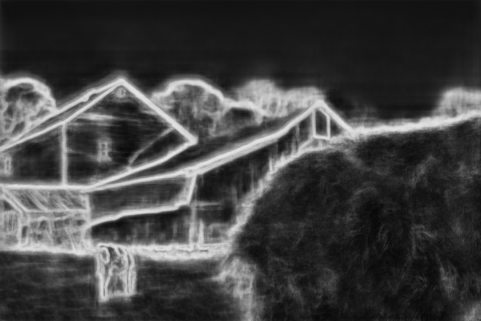}&
\hspace{\ll}\roundedpicture[width = .16\textwidth]{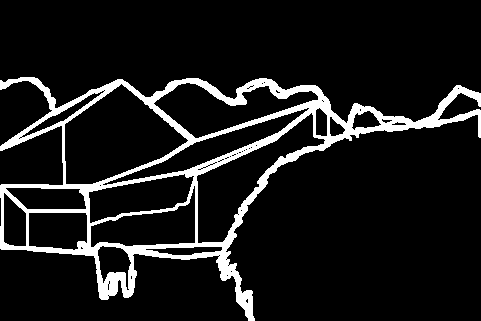}\\[-.5ex]
\hspace{\lll}Input Image&\hspace{\ll}CHM result &\hspace{\ll} Groundtruth\\
\end{tabular}
\caption{Results of CHM on different tasks.
\textbf{First row: }Scene labeling (Stanford background dataset~\cite{Gould2009}.
\textbf{Second row: }Horse segmentation (Weizmann dataset~\cite{Borenstein2004}.
\textbf{Third row: } Membrane detection (mouse neuropil dataset~\cite{SeyedhosseiniMulti2013}).
\textbf{Fourth row: }Edge detection (Berkeley dataset~\cite{Martin2001}).
See section~\ref{Sec:exp} for details.}
\label{Fig:samples}
\end{figure}

In extensive experiments, we demonstrate the performance of CHM on a couple of challenging vision tasks: Horse segmentation in the Weizmann dataset~\cite{Borenstein2004}, outdoor scene labeling in the Stanford background~\cite{Gould2009}.
We also show the performance of CHM for edge detection on the popular BSDS 500~\cite{Martin2001} and NYU Depth (v2)~\cite{Silberman2011} datasets.
In all cases, CHM results in either state-of-the-art or near state-of-the-art performance.
In addition, we apply CHM on two electron microscopy datasets for cell membrane detection (Drosophila VNC~\cite{Cardona2010,Cardona2012} and mouse neuropil~\cite{SeyedhosseiniMulti2013}).
CHM outperforms many existing algorithms for membrane detection and can be used as the first step towards reconstruction of the connectome, \textit{i.e.,} the map of neural connectivity in the mammalian nervous system~\cite{Sporns2005}. Some samples of CHM results are shown in Figure~\ref{Fig:samples}. 

An early version of this work was first presented in~\cite{SeyedhosseiniImage2013}. This journal version reports more comprehensive experiments and gives more theoretical insight into CHM.
\section{Related Work}
\subsection{Graphical Models}
There have been many methods that employ graphical models to take advantage of contextual information for scene labeling.
Markov Random Fields (MRF)~\cite{Larlus2008Combining,Kumar2010Efficiently,Gould2009,Tighe2010Superparsing} and Conditional Random Fields (CRF)~\cite{He2004,Ladicky2009Associative} are the most popular approaches.
He \textit{et al.}~\cite{He2004} used CRF to capture contextual information at multiple scales.
Larlus and Jurie~\cite{Larlus2008Combining} used MRF on top of a bag-of-words based object model to ensure consistency of labeling.
Gould \textit{et al.}~\cite{Gould2009} defined an energy function over scene appearance and geometry and then developed an efficient inference technique for MRFs to minimize that energy.
Kumar and Koller~\cite{Kumar2010Efficiently} formulated the energy minimization as an integer programming problem and proposed a linear programming relaxation to solve it.
Tighe and Lazebnik~\cite{Tighe2010Superparsing} proposed an MRF-based superpixel matching that can be easily scaled to large datasets.
Ladicky \textit{et al.}~\cite{Ladicky2009Associative} introduced a hierarchical CRF, which is able to combine features extracted from pixels and segments.
For inference, they used a graph-cut~\cite{Boykov2004Experimental} based method to find the MAP solution.
Ren \textit{et al.}~\cite{RenRGB2012} used a superpixel MRF together with a segmentation tree for RGB-D scene labeling.

Many of graphical methods rely on presegmentation to superpixels~\cite{Tighe2010Superparsing,RenRGB2012} or multiple segment candidates~\cite{Kohli2009Robust,Kumar2010Efficiently}.
More powerful region-based features can be extracted from superpixels compared to pixels. Moreover, presegmentation to superpixels improves the computational efficiency of these models.
However, it is known that superpixels might not adhere to the image boundaries~\cite{Achanta2012Slic} and thus can decrease labeling accuracy~\cite{RenRGB2012}.
This motivated approaches using multiple segments as hypothesis.
However, these methods can be problematic when dealing with cluttered images~\cite{Ladicky2009Associative}.
This motivated methods with hierarchical segmentation~\cite{Ladicky2009Associative,Munoz2010Stacked}.

Unlike previously cited approaches, our proposed method does not make use of any presegmentations or exemplars and works directly on image pixels. This allows our model to be applied to different problems without any modifications.
Moreover, inference is simpler in our CHM compared to graphical models. It only requires the evaluation of classifier function and does not require searching the label space as in CRFs~\cite{Grangier2009Deep}.

\subsection{Convolutional Networks}
Deep learning is a very active area of research and has been widely used in the computer vision field.
Convolutional networks (ConvNet)~\cite{Lecun1998Gradient} are one of the most popular deep architectures.
They were initially proposed for character recognition~\cite{Lecun1998Gradient}, but later applied successfully to image classification~\cite{Krizhevsky2012Imagenet,Ciresan2012Multi} and object detection~\cite{Szegedy2013Deep,Jarrett2009What}.
They have also been used for biological image segmentation~\cite{Jain2007,Turaga2009Maximin,Ciresan2012Deep} and scene labeling~\cite{Grangier2009Deep,Farabet2013}.
Jain~\textit{et al.} used ConvNets to restore membranes in electron microscopic (EM) images.
Turaga \textit{et al.}~\cite{Turaga2009Maximin} used ConvNets to minimize the Rand index~\cite{Rand1971} instead of pixel error to improve the segmentation of EM images.
Ciresan \textit{et al.} trained a very large ConvNet with four convolutional layers followed by two fully connected layers.
This method was used in the winning entry of the ISBI neuronal segmentation challenge~\cite{ISBI2012}.
Grangier \textit{et al.}~\cite{Grangier2009Deep} trained a ConvNet by iteratively adding new layers for scene parsing.
Farabet \textit{et al.}~\cite{Farabet2013} proposed a multi-scale ConvNet for scene parsing.
Their framework contains multiple copies of a single network which are applied to a scale-space pyramid of input images.
They performed some postprocessing methods to clean up the outputs generated by the ConvNet.

ConvNets can cover large contextual area compared to other methods, but they need several hidden layers with many free parameters.
Training the ConvNets is computationally expensive and might take months or even years on CPUs~\cite{Ciresan2012Deep}. Hence, GPU implementations, which speed up the training process, are usually needed in practice.
Unlike ConvNets, our CHM can be trained on CPUs in a reasonable time. Moreover, we will show that CHM outperforms the ConvNets proposed in~\cite{Jain2007,Ciresan2012Deep,Farabet2013}.

\subsection{Cascaded Classifiers}
The idea of using multiple classifiers to model context has been proven successful to solve different computer vision problems.
Fink and Perona~\cite{Fink2004} proposed the mutual boosting framework which takes advantage of multiple detectors in a boosting architecture for object detection.
Torralba \textit{et al.}~\cite{Torralba2004} proposed the boosted random field (BRF), which uses boosting to learn the graph structure of CRFs, for object detection and segmentation.
Heitz \textit{et al.}~\cite{Heitz2008} proposed a different architecture to combine multiple classifiers, called cascaded classifier model, for holistic scene understanding.
Their model combines several classifiers tuned for some specific subtasks to improve the performance on all subtasks.
Li~\textit{et al.}~\cite{Li2012} introduced a feedback enabled cascaded classification model which jointly optimizes several subtasks in a two-layer cascade of classifiers.
In a more related work, Tu and Bai~\cite{Tu2010} introduced the auto-context algorithm, which integrates both image features and contextual information to learn a series of classifiers, for image segmentation.
A filter bank is used to extract the image features and the output of each classifier is used as the contextual information for the next classifier in the series. 
Jurrus \textit{et al.}~\cite{Jurrus2010} also trained a series of artificial neural networks (ANN)~\cite{Haykin1999}, which learns a set of convolutional filters from the data instead of applying fixed filter banks to the input image.
Their series architecture was improved by employing a multi-scale representation of context during training~\cite{Seyedhosseini2011}.
The advantage of the cascaded classifier model over ConvNets is its easier training due to treating each classifier in the series one at a time.

We also introduce a segmentation framework that takes advantage of both input image features and contextual information. Similar to the auto-context algorithm, we use a filter bank to extract input image features. But we use a hierarchical architecture to capture contextual information at different resolutions.
Moreover, this multi-resolution contextual information is learned in a supervised framework, which makes it more discriminative compared to the above-mentioned methods.
From the Bayesian point of view, CHM optimizes a joint posterior probability at multiple resolutions simultaneously.   
To our knowledge, supervised multi-resolution contextual information has not previously been used in a scene labeling framework.
\subsection{Edge Detection}
There is a large body of work in the area of edge detection.
Many unsupervised techniques have been proposed for edge detection~\cite{Canny1986,Perona1990,Arbelaez2009,Arbelaez2011Contour}.
Seminal Canny edge detector~\cite{Canny1986} is one of the earliest and gPb~\cite{Arbelaez2011Contour} is one of the latest among these approaches.
More recently, supervised techniques have been explored to improve the edge detection performance~\cite{Martin2004Learning,Dollar2006,Mairal2008Discriminative,Ren2012,Lim2013Sketch,Dollar2013}.
Martin \textit{et al.}~\cite{Martin2004Learning} computed gradients for brightness, color, and texture channels on a circular disc located at each pixel.
They then combined these hand-crafted features and used them as input to a logistic regression classifier for predicting edges.
Dollar \textit{et al.}~\cite{Dollar2006} extracted tens of thousands of features at each pixel, and then used a probabilistic boosting tree (PBT)~\cite{Tu2005} to find edges.
Mairal \textit{et al.}~\cite{Mairal2008Discriminative} proposed to learn discriminative sparse dictionaries to distinguish between ``patches centered on an edge pixel'' and ``patches centered on a non-edge pixel''.
Ren and Bo~\cite{Ren2012} used gradients over learned sparse codes instead of hand designed gradients of~\cite{Martin2004Learning} to achieve state-of-the-art performance.
Lim \textit{et al.}~\cite{Lim2013Sketch} defined a set of sketch tokens by clustering the patches extracted from groundtruth images. Then, they trained a random forest to detect those tokens at test time.
Finally, Dollar and Zitnick~\cite{Dollar2013} made use of different edge patterns, \textit{e.g.,} T-junctions and Y-junctions, present in images, and used a structured random forest to learn those patterns. Their method is fast and generalizes well between different datasets.

We also approach the edge detection problem as a labeling problem. Our CHM is trained to distinguish between ``patches centered on an edge pixel'' and ``patches centered on a non-edge pixel''. We will show that CHM achieves near state-of-the-art performance on the Berkeley dataset~\cite{Martin2001} and outperforms state-of-the-art methods~\cite{Ren2012,Dollar2013} on NYU depth dataset.
Moreover, we will demonstrate that generalization performance of CHM across different datasets is better compared to~\cite{Ren2012,Dollar2013}.

%
%
%
%
%

\section{Contextual Hierarchical Model}
\begin{figure}
\includegraphics[width = \linewidth ,height = 9 cm]{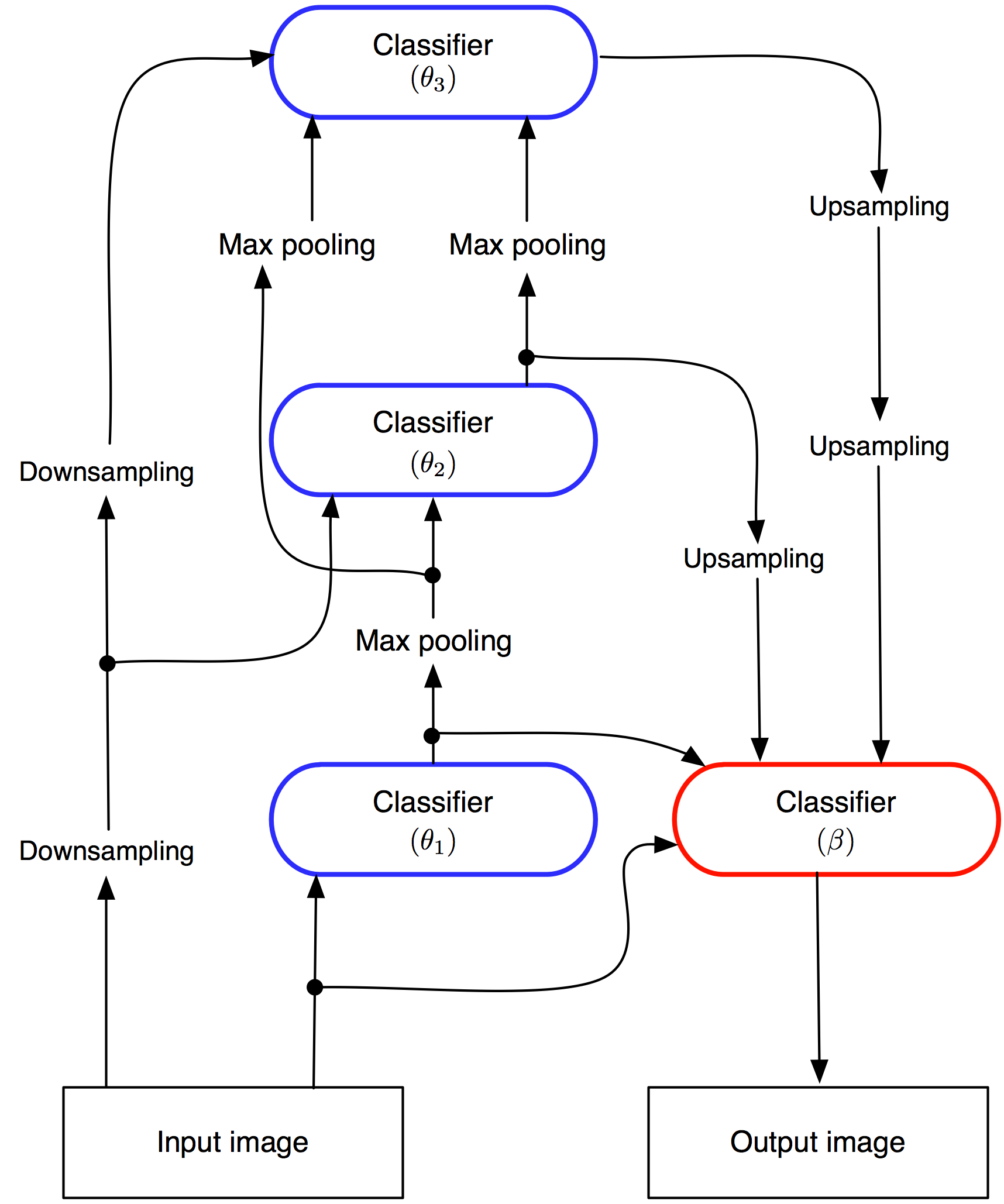}
\caption{Illustration of the contextual hierarchical model. The blue classifiers are learned during the bottom-up step and the red classifier is learned during the top-down step. In the bottom-up step, each classifier takes the outputs of lower classifiers as well as the input image as input. The height of the hierarchy, L, is three in this model but it can be extended to any arbitrary number.}
\label{Fig:CHM}
\end{figure}

The contextual hierarchical model (CHM) is illustrated in Figure~\ref{Fig:CHM}.
First, a multi-resolution representation of the input image is obtained by applying downsampling sequentially. 
Next, a series of classifiers are trained at different resolutions from the finest resolution to the coarsest resolution.
At each resolution, the classifier is trained based on the outputs of the previous classifiers in the hierarchy and the input image at that resolution.
Finally, the outputs of these classifiers are used to train a new classifier at original resolution.
This classifier exploits the rich contextual information from multiple resolutions.
The whole training process targets a joint posterior probability at multiple resolutions (see section~\ref{Sec:Prob}).
We describe different steps of the model separately in the following subsections.

\subsection{Bottom-up step}
Let $X = (x(m,n))$ be the $2D$ input image with a corresponding ground truth $Y = (y(m,n))$ where $y(m,n) \in \{0,1\}$ is the class label for pixel $(m,n)$.
For notational simplicity, we use $1D$ vectors $X = (x_1,x_2,\dots,x_n)$ and $Y = (y_1,y_2,\dots,y_n)$ to denote the input image and corresponding ground truth, respectively.
The training dataset then contains $K$ input images, $\mathbf{X} = \{X_1,X_2,\dots,X_K\}$, and corresponding ground truth images, $\mathbf{Y} = \{Y_1,Y_2,\dots,Y_K\}$\footnote{Unless specified otherwise, upper case symbols, \textit{e.g.,} $X$, $Y$, denote a particular vector, lower case symbols, \textit{e.g.,} $x$, $y$, denote the elements of a vector, and bold-face symbols, \textit{e.g.,} $\mathbf{X}$ , $\mathbf{Y}$, denote a set of vectors.}.
We also define the $\Phi(\cdot,l)$ operator which performs down-sampling $l$ times by averaging the pixels in each $2\times 2$ window, and the $\Gamma(\cdot,l)$ operator which performs max-pooling $l$ times by finding the maximum pixel value in each $2\times 2$ window.
Each classifier in the hierarchy has some internal parameters $\theta_l$, which are learned during training 
\begin{align}
\nonumber \mathbf{\hat{\theta_l}} = \argmax_{\mathbf{\theta_l}} P(&\mathbf{\Gamma}(\mathbf{Y},l-1)\mid \mathbf{\Phi}(\mathbf{X},l-1),\\ &\mathbf{\Gamma(\hat Y^{1}},l-1),\dots, \mathbf{\Gamma(\hat Y^{l-1}},1);\mathbf{\theta_l})
\label{Eq:LearnTD}
\end{align}
where $\mathbf{\hat Y^{1}},\dots,\mathbf{\hat Y^{l-1}}$ are the outputs of classifiers at the lower levels of the hierarchy. The classifier output of each level is obtained using inference 
\begin{align}
 \nonumber \hat Y^l = \argmax_{Y}P(&Y \mid \Phi(X,l-1),\\ &\Gamma({\hat Y^{1},l-1}),\dots,\Gamma({\hat Y^{l-1},1});\mathbf{\hat{\theta_l}}).
\label{Eq:InferTD}
\end{align}
Each classifier in the $l$'th level of the hierarchy takes outputs of all lower level classifiers, i.e., $\hat{Y^1}, \dots, \hat{Y}^{l-1}$, which provide multi-resolution contextual information.
For $l=1$ no prior information is used and the classifier parameters, $\theta_1$, are learned only based on the input image.

It is worth mentioning that classifiers at higher levels of the hierarchy have access to contextual information from larger areas because they are trained on downsampled images.
\subsection{Top-down step}
Unlike the bottom-up step where multiple classifiers are learned, only one classifier is trained in the top-down step.
Once all the classifiers are learned in the bottom-up step, a top-down path is used to feed coarser resolution contextual information into a classifier, which is trained at the finest resolution.
We define $\Omega(\cdot,l)$ operator that performs  upsampling $l$ times by duplicating each pixel. For a hierarchical model with $L$ levels, the classifier is trained based on the input image and the outputs of stages $1$ to $L$ obtained in the bottom-up step. The internal parameters of the classifier, $\mathbf{\beta}$, are learned using the following
\begin{align}
\nonumber \small \mathbf{\hat \beta} = \argmax_{\mathbf{\beta}} P(\mathbf{Y}\mid \mathbf{X},\mathbf{\hat Y^{1}},\mathbf{\Omega(\hat Y^{2}},1),\dots,\\ \mathbf{\Omega(\hat Y^{L}},L-1);\mathbf{\beta}).
\label{Eq:LearnBU}
\end{align}
The output of this classifier can be obtained using the following for inference
\begin{align}
\nonumber \small \hat Z = \argmax_{Y}P(Y \mid X,\hat Y^{1},\Omega(\hat Y^{2},1),\dots,\\ \Omega(\hat Y^{L},L-1);\mathbf{\hat{\beta}}).
\label{Eq:InferBU}
\end{align}
The top-down classifier takes advantage of prior information from multiple resolutions.
This multi-resolution prior is an efficient mixture of both local and global information since it is drawn from different scales.
In a related work, Seyedhosseini \textit{et al.}~\cite{Seyedhosseini2011} proposed a multi-scale contextual model that exploits contextual information from multiple scales. The advantage of the model proposed here is that the context images are learned at different scales in a supervised framework while the multi-scale contextual model uses simple filtering to create context images at different scales.
This allows CHM to optimize a joint posterior at different scales.
The overall learning and inference algorithms for the contextual hierarchical model are described in Algorithm~\ref{AL:CHM} and Algorithm~\ref{AL:CHMinfer}, respectively.
\begin{algorithm}                    
\caption{Learning algorithm for the CHM.}          
\label{AL:CHM}                           
\begin{algorithmic}                    
    
    \Require A set of training images together with their binary groundtruth images, $\mathbf{S} = \{ (X_i,Y_{i}), i=1,\dots,K\}$ and the height of the hierarchy, $L$.
    
    \Ensure $\mathbf{\Theta} = \{ \hat\theta_{1}, \dots , \hat\theta_{L}, \hat\beta \}$.
    
    \begin{itemize}
    \item {Learn the first classifier, $\theta_{1}$, using equation~(\ref{Eq:LearnTD}) without any prior information and only based on the input image features.}
    \item {Compute the output of first classifier, $\mathbf{\hat Y^{1}}$, using equation~(\ref{Eq:InferTD}).}
	\end{itemize}  	

		\For{$l = 2\ to\ L$}
		\begin{itemize}	 
		\setlength{\itemindent}{\algorithmicindent}
			 \item Learn the $l$'th classifier, $\mathbf{\hat \theta_{l}}$, using equation~(\ref{Eq:LearnTD}).
			\item Compute output of the $l$'th classifier, $\mathbf{\hat Y^{l}}$, using equation~(\ref{Eq:InferTD}).
		\end{itemize}
		\EndFor	
		\begin{itemize}
		\item Learn the top-down classifier, $\mathbf{\hat \beta}$, using equation~\ref{Eq:LearnBU}.
		\end{itemize}

\end{algorithmic}
\end{algorithm}
\begin{algorithm}                    
\caption{Inference algorithm for the CHM.}          
\label{AL:CHMinfer}                           
\begin{algorithmic}                    
    
    \Require An input image $X$, $\mathbf{\Theta}$, $L$.
    
    \Ensure $\hat Z$.
    
    \begin{itemize}
    \item {Compute the output of first classifier, ${\hat Y^{1}}$, using equation~(\ref{Eq:InferTD}).}
	\end{itemize}  	

		\For{$l = 2\ to\ L$}
		\begin{itemize}	 
		\setlength{\itemindent}{\algorithmicindent}
			\item Compute output of the $l$'th bottom-up classifier, ${\hat Y^{l}}$, using equation~(\ref{Eq:InferTD}).
		\end{itemize}
		\EndFor	
		\begin{itemize}	 
		\setlength{\itemindent}{\algorithmicindent}
			\item Compute output of the top-down classifier, ${\hat Z}$, using equation~(\ref{Eq:InferBU}).
		\end{itemize}		
\end{algorithmic}
\end{algorithm}
\subsection{Probabilistic Interpretation} \label{Sec:Prob}
Given the training set $\mathbf{X}$, containing $T = K \times n$ samples, and corresponding labels $\mathbf{Y}$, a common approach is to find the optimal solution by solving the maximum aposteriori (MAP) equation
\begin{equation}
\log \prod_t P(Y_t \mid X_t;\mathbf{\Theta}).
\end{equation}
There are two common strategies to solve this optimization.
The first strategy, \textit{i.e.,} generative approach, decomposes the posterior to likelihood, $P(X_t \mid Y_t)$, and prior, $P(Y_t)$.
The second strategy, \textit{i.e.,} discriminative approach, targets the posterior distribution directly.
Our hierarchical model falls into the second category.
However, it differs from other approaches in a sense that it optimizes a joint posterior at multiple resolutions, \textit{i.e.,}
\begin{align}
\label{Eq:MMAP}
\nonumber \log \prod_t  P\left(Y_t,\Gamma(Y_t,0),\dots,\Gamma(Y_t,L-1) \mid X_t;\mathbf{\Theta} \right) = \\
\sum_t \log P\left( Y_t, \Gamma(Y_t,0),\dots,\Gamma(Y_t,L-1) \mid X_t;\mathbf{\Theta} \right),
\end{align}
where $\Gamma$ is the maxpooling operator and $L$ is the number of levels in the hierarchy.
Using $P(A,B \mid C) = P(A \mid B,C) P(B \mid C)$, equation~\ref{Eq:MMAP} can be rewritten as
\begin{align}
\nonumber \sum_t &\log \Big( P\big(Y_t \mid X_t,\Gamma(Y_t,0),\dots,\Gamma(Y_t,L-1);\mathbf{\Theta} \big) \times \\ \nonumber &P\big(\Gamma(Y_t,L-1) \mid X_t,\Gamma(Y_t,0),\dots,\Gamma(Y_t,L-2);\mathbf{\Theta} \big) \times \\
\nonumber &\dots \times P\big( \Gamma(Y_t,0) \mid X_t;\mathbf{\Theta}\big) \Big)=
\end{align}
\begin{align}
\nonumber &\underbrace{\sum_t \log  P\big(Y_t \mid X_t,\Gamma(Y_t,0),\dots,\Gamma(Y_t,L-1);\mathbf{\Theta} \big) }_{\text{Top-down:}J_2(\mathbf{X},\mathbf{Y};\mathbf{\Theta})} + 
\\
&\underbrace{\sum_l \sum_t \log P\big(\Gamma(Y_t,l) \mid X_t, \Gamma(Y_t,0),\dots,\Gamma(Y_t,l-1);\mathbf{\Theta} \big) }_{\text{Bottom-up:}J_1(\mathbf{X},\mathbf{Y};\mathbf{\Theta})}.
\end{align}
Note that the optimization problems nicely splits down to two subproblems, \textit{i.e.,} $J_1(\mathbf{X},\mathbf{Y};\mathbf{\Theta})$ and $J_2(\mathbf{X},\mathbf{Y};\mathbf{\Theta})$, which are solved during bottom-up and top-down steps respectively.

In practice, the optimization is done in a greedy way.
The output of the classifier at level $l$, $\hat Y^{l}$, is used as an approximation of the groundtruth at that resolution, $\Gamma(Y,l-1)$.
Therefore, the following optimization problems are solved during training
\begin{align}
\nonumber &\textbf{Bottom-up:}\\ \nonumber &\max_\mathbf{\Theta} J_1(\mathbf{X},\mathbf{Y};\mathbf{\Theta}) = \\ &\max_\mathbf{\Theta} \sum_l \sum_t \log P\big(\Gamma(Y_t,l) \mid X_t, \hat Y_{t}^{1},\dots,\hat Y_{t}^{l});\mathbf{\Theta} \big)\\
\nonumber &\textbf{Top-down:}\\
\nonumber &\max_\mathbf{\Theta} J_2(\mathbf{X},\mathbf{Y};\mathbf{\Theta}) =\\
&\max_\mathbf{\Theta} \sum_t \log  P\big(Y_t \mid X_t,\hat Y_t^1,\dots,\hat Y_t^L;\mathbf{\Theta} \big) 
\end{align}
This greedy approach makes the training simple and tractable.
It is noteworthy that each of the terms of the outer summation in $J_1$ is corresponding to one level of the hierarchy.
Due to the greedy optimization, a second stage of CHM can improve the results.
In the second stage, the top-down classifier of the previous stage is used as the first classifier in the bottom-up step.

%

\subsection{Classifier selection}
Even though our problem formulation is general and not restricted to any specific type of classifier, in practice we need a fast and accurate classifier that is robust against overfitting.
Among off-the-shelf classifiers, we consider artificial neural networks (ANN), support vector machines (SVM), and random forests (RF).
ANNs are slow at training time due to the computational cost of backpropagation.
SVMs offer good generalization performance, but choosing the kernel function and the kernel parameters can be time consuming since they need to be adopted for each classifier in the CHM.
Furthermore, SVMs are not intrinsically probabilistic and thus are not completely suitable for our CHM model.
Random forests provide an unbiased estimate of testing error, but they are prone to overfitting in the presence of noise.
In section~\ref{Sec:weizmann} we show that overfitting can disrupt learning in the CHM model.

We adopt logistic disjunctive normal networks (LDNN)~\cite{SeyedhosseiniImage2013} as the classifier in CHM.
LDNN is a powerful classifier, which consists of one adaptive layer implemented by logistic sigmoid functions followed by two fixed layers of logical units that compute conjunctions and disjunctions, respectively.
LDNN allows an intuitive initialization using k-means clustering and outperforms neural networks, SVMs, and random forests on several standard datasets~\cite{SeyedhosseiniImage2013}.
Finally, LDNNs are fast to train due to the single adaptive layer, which makes them suitable for the CHM architecture.

\subsection{Feature selection} \label{Sec:features}
In this section, we describe the set of features extracted from input and context images in CHM.
The features that we extract from input images include Haar features~\cite{Viola2004} and histogram of oriented gradients (HOG) features~\cite{Dalal2005}.
These features are efficient to compute and somewhat complementary to each other~\cite{Tu2010}.
For color images, Haar and HOG features are computed for each channel separately.
We also use dense SIFT flow features~\cite{Liu2011sift} computed at each pixel.
In addition, we apply a set of Gabor filters with different parameters and Canny edge detector to obtain more features.
Beside these appearance features, we also use position and its higher moments (up to $2^{nd}$ order), which are known to be informative for scene labeling~\cite{RenRGB2012,Munoz2010Stacked}.
Finally, we use a $15\times 15$ sparse stencil structure, which contains $57$ samples, to sample the neighborhood around each pixel.
In summary, we extract $647$ features from color images and $457$ features from gray scale images.

Context features are obtained from the outputs of classifiers in the hierarchy.
We used a $15 \times 15$ stencil to sample context images around each pixel.
We also tried larger and more dense sampling structures, \textit{e.g.,} $21 \times 21$ patch, but they had negligible impact on the performance.
We do not extract any other features beside the neighborhood samples from context images.

\section{Experimental Results}\label{Sec:exp}
We perform experimental studies to evaluate the performance of CHM on three different applications: Scene labeling, edge detection, and biomedical image segmentation.
The diversity among these applications shows the broad applicability of our method.
In all the applications, we used a set of nearly identical parameters, including the number of levels in CHM and the features parameters.
Following the reproducible research instructions~\cite{RR}, we maintain a web page containing the source codes and scripts used to generate the results in this section\footnote{\url{http://www.sci.utah.edu/~mseyed/Mojtaba_Seyedhosseini/CHM.html}}.

\subsection{Scene Labeling}
We show the performance of CHM on a binary scene labeling dataset, \textit{i.e.,} Weizmann dataset~\cite{Borenstein2004}, as well as an outdoor scene labeling dataset with multiple classes, \textit{i.e.,} Stanford background dataset~\cite{Gould2009}.

\subsubsection{Weizmann dataset} \label{Sec:weizmann}
The Weizmann dataset~\cite{Borenstein2004} contains $328$ gray scale horse images with corresponding foreground/background truth maps.
Similar to Tu \textit{et al.}~\cite{Tu2010}, we used half of the images for training and the remaining images were used for testing.
The task is to segment horses in each image.
We used the features described in section~\ref{Sec:features}. Note that we do not use location information for this dataset since horses are mostly centered in the images, which would create an unfair advantage.

We used a $24 \times 24$ LDNN as the classifier in a CHM with two stages and $5$ levels per stage.
To improve the generalization performance, we adopted the dropout idea.
Hinton \textit{et al.}~\cite{Hinton2012} showed that removing $50\%$ of the hidden nodes in a neural network during the training can improve the performance on the test data.
Using the same idea, we randomly removed half of the nodes in the second layer and half of the nodes per group in the first layer at each iteration during the training.
At test time, we used the LDNN that contains all of the nodes with their outputs square rooted to compensate for the fact that half of them were active during the training time.

For comparison, we trained a CHM with random forest as the classifier. To avoid overfitting, only $\frac{1}{20}$ of samples were used to train $100$ trees in the random forest. We also trained a multi-scale series of artificial neural networks (MSANN) as in~\cite{Seyedhosseini2011}. Three metrics were used to evaluate the segmentation accuracy: Pixel accuracy, F-value $= \frac{2\times precision \times recall}{precision+recall}$, and G-mean$= \sqrt{recall\times TNR}$ where $TNR = \frac{true\ negative}{true\ negative+false\ positive}$. Unlike F-value, G-mean is symmetric with respect to positive and negative classes.
In Table~\ref{Tab:Horse} we compare the performance of CHM with some state-of-the-art methods.
\begin{table}
\caption{Testing performance of different methods on the Weizmann horse dataset.}
\centering
\begin{tabular}{cccc} \toprule
Method & \small F-value & \small G-mean & \small Pixel accuracy  \\ \midrule
KSSVM~\cite{Bertelli2011}&$?$&$?$&$94.60\%$\\ \midrule
TWM~\cite{Kuettel2012}&$?$&$?$&$94.70\%$\\ \midrule
\small Auto-context~\cite{Tu2010}&$84\%$&$?$&$?$\\ \midrule
\small Levin \& Weiss~\cite{Levin2006}&$?$&$?$&$95.2\%$\\ \midrule
MSANN~\cite{Seyedhosseini2011}&$87.58\%$&$92.76\%$&$94.34\%$\\ \midrule
CHM-RF&$83.15\%$&$90.20\%$&$92.33\%$\\ \midrule
CHM-LDNN&$\bf{89.89}\%$&$\bf{94.39}\%$&$\bf{95.37}\%$\\ \bottomrule
\end{tabular}
\label{Tab:Horse}
\end{table}
CHM outperforms other state-of-the-art methods. It is worth noting that CHM does not make use of fragments and it is based purely on discriminative classifiers that use neighborhood information.

\begin{figure}
\centering
\includegraphics[width = .7\linewidth]{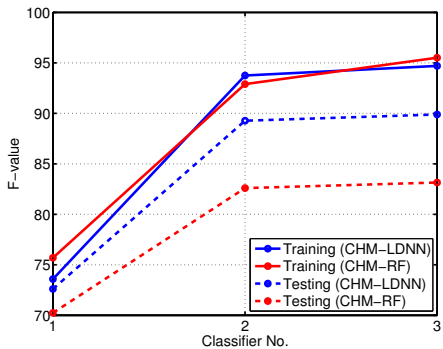}
\caption{
F-value of the classifiers trained at the original resolution in the CHM with LDNN and random forest. The overfitting in the random forest makes it useless in the CHM architecture.
}
\label{Fig:CHMStage}
\end{figure}

\begin{figure}
\def \wii{.226\linewidth}
\def \hii{.15\linewidth}
\begin{tabular}{ccccc}
\hspace{-.15cm}\begin{sideways}\fontsize{9pt}{0}\fontfamily{phv}\selectfont \hspace*{.3cm} (a)\end{sideways}&\hspace{-4mm}\includegraphics[width = \wii, height = \hii]{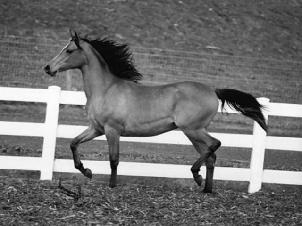}\hspace{-3mm} & \includegraphics[width = \wii, height = \hii]{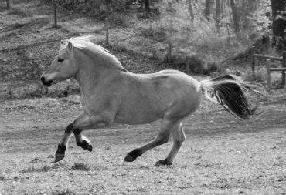}\hspace{-3mm} & \includegraphics[width = \wii, height = \hii]{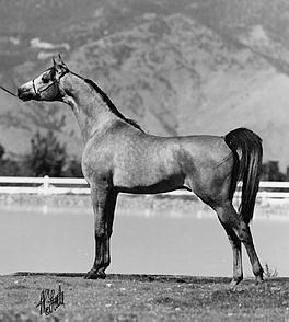}\hspace{-3mm} & \includegraphics[width = \wii, height = \hii]{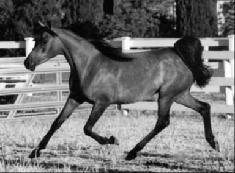}\hspace{-3mm} \\

\hspace{-.15cm}\begin{sideways}\fontsize{9pt}{0}\fontfamily{phv}\selectfont \hspace*{.3cm} (b)\end{sideways}&\hspace{-4mm}\includegraphics[width = \wii, height = \hii]{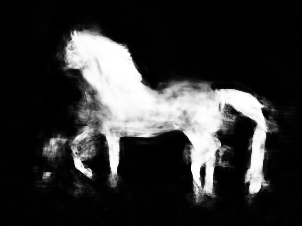}\hspace{-3mm} & \includegraphics[width = \wii, height = \hii]{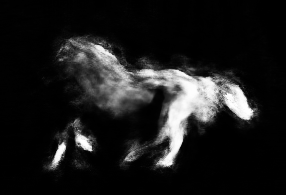}\hspace{-3mm} & \includegraphics[width = \wii, height = \hii]{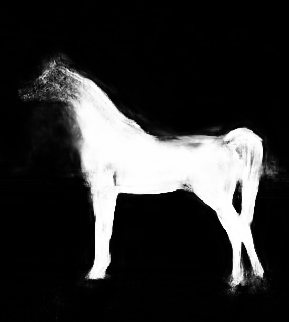}\hspace{-3mm} & \includegraphics[width = \wii, height = \hii]{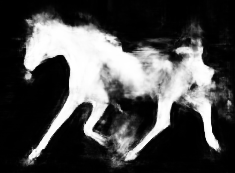}\hspace{-3mm} \\

\hspace{-.15cm}\begin{sideways}\fontsize{9pt}{0}\fontfamily{phv}\selectfont \hspace*{.3cm} (c)\end{sideways}&\hspace{-4mm}\includegraphics[width = \wii, height = \hii]{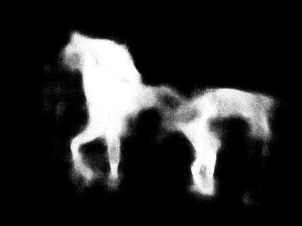}\hspace{-3mm} & \includegraphics[width = \wii, height = \hii]{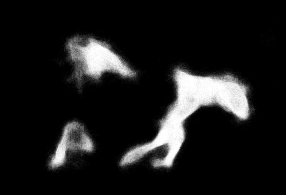}\hspace{-3mm} & \includegraphics[width = \wii, height = \hii]{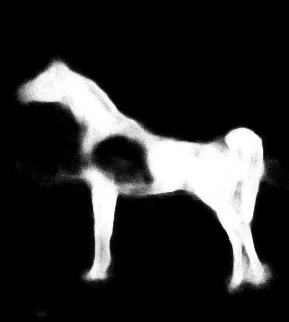}\hspace{-3mm} & \includegraphics[width = \wii, height = \hii]{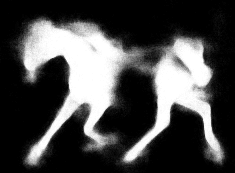}\hspace{-3mm} \\

\hspace{-.15cm}\begin{sideways}\fontsize{9pt}{0}\fontfamily{phv}\selectfont \hspace*{.3cm} (d)\end{sideways}&\hspace{-4mm}\includegraphics[width = \wii, height = \hii]{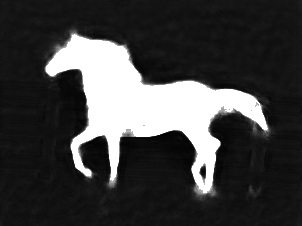}\hspace{-3mm} & \includegraphics[width = \wii, height = \hii]{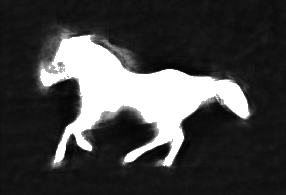}\hspace{-3mm} & \includegraphics[width = \wii, height = \hii]{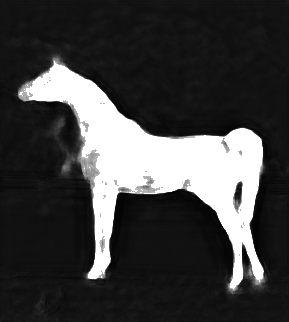}\hspace{-3mm} & \includegraphics[width = \wii, height = \hii]{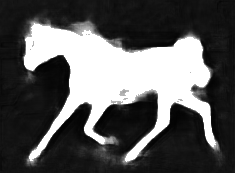}\hspace{-3mm} \\

\hspace{-.15cm}\begin{sideways}\fontsize{9pt}{0}\fontfamily{phv}\selectfont \hspace*{.3cm} (e)\end{sideways}&\hspace{-4mm}\includegraphics[width = \wii, height = \hii]{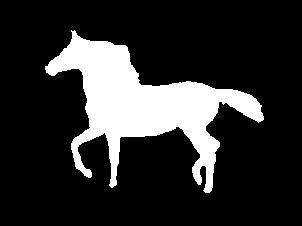}\hspace{-3mm} & \includegraphics[width = \wii, height = \hii]{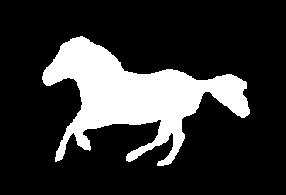}\hspace{-3mm} & \includegraphics[width = \wii, height = \hii]{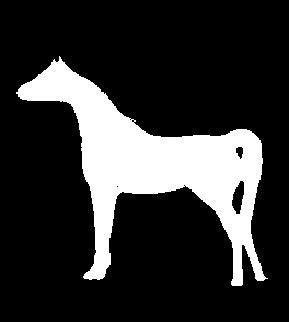}\hspace{-3mm} & \includegraphics[width = \wii, height = \hii]{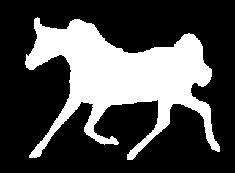}\hspace{-3mm} \\

\end{tabular}
\caption{Test results of the Weizmann horse dataset. (a) Input image, (b) MSANN~\cite{Seyedhosseini2011}, (c) CHM-RF, (d) CHM-LDNN, (e) ground truth images. The CHM-LDNN is more successful in completing the body of horses.}
\label{Fig:Horse}
\end{figure}

The CHM-LDNN outperforms the state-of-the-art methods while the CHM-RF performs worse than those methods.
The training and testing F-value of the classifiers trained at the original resolution in the CHM, \textit{i.e.,} the classifiers at the bottom of hierarchy, for both LDNN and random forest are shown in Figure~\ref{Fig:CHMStage}.
It shows how overfitting propagates through the stages of the CHM when the random forest is used as the classifier. The overfitting disrupts the learning process because there are too few mistakes in the training set compared to the testing set as we go through the stages. For example, the overfitting in the first stage does not permit the second stage to learn the typical mistakes from the first stage that will be encountered at testing time. We tried random forests with different parameters to overcome this problem but were unsuccessful.
Figure~\ref{Fig:Horse} shows four examples of our test images and their segmentation results using different methods. The CHM-LDNN outperforms the other methods in filling the body of horses.

\subsubsection{Stanford background dataset}
The Stanford background dataset~\cite{Gould2009} contains $715$ images of urban and rural scenes, collected from other public datasets such that each image is approximately $240 \times 320$ pixels and contains at least one foreground object.
This dataset is composed of eight classes, one foreground and seven other classes, and the groundtruth images, obtained from Amazon Mechanical Turk, are included in the dataset.
We followed the standard evaluation procedure for this dataset, which is performing $5$-fold cross-validation with the dataset randomly split into $572$ training images and $143$ test images.

\begin{figure*}
\def \ll{-.45cm}
\def \ww{0.15\textwidth}
\def \hh{0.11\textwidth}
\centering
\begin{tabular}{cccccc}
\roundedpicture[width = \ww, height = \hh]{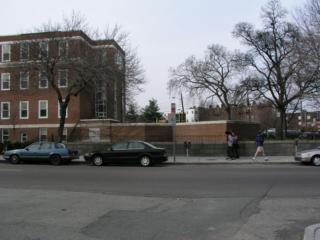}&\hspace{\ll}
\roundedpicture[width = \ww, height = \hh]{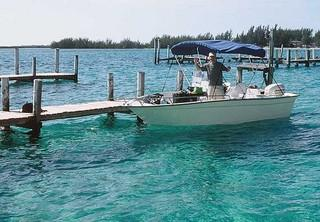}&\hspace{\ll}
\roundedpicture[width = \ww, height = \hh]{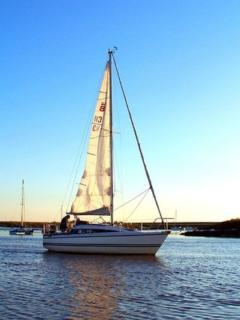}&\hspace{\ll}
\roundedpicture[width = \ww, height = \hh]{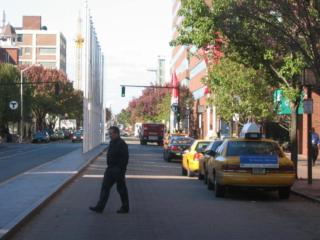}&\hspace{\ll}
\roundedpicture[width = \ww, height = \hh]{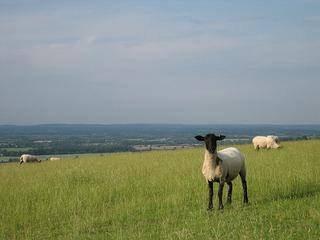}&\hspace{\ll}
\roundedpicture[width = \ww, height = \hh]{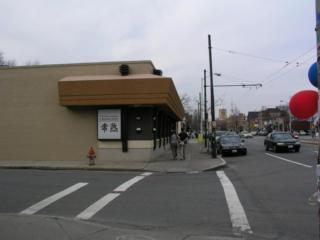}\\

\roundedpicture[width = \ww, height = \hh]{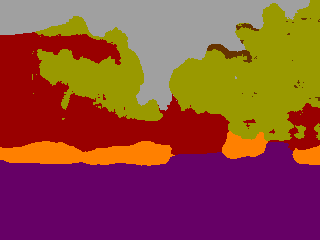}&\hspace{\ll}
\roundedpicture[width = \ww, height = \hh]{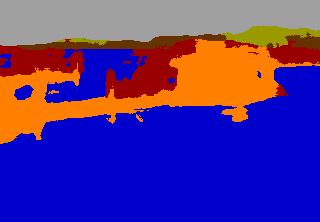}&\hspace{\ll}
\roundedpicture[width = \ww, height = \hh]{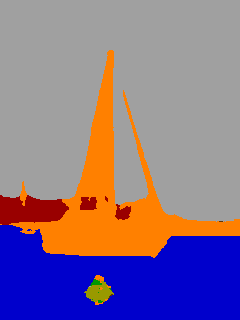}&\hspace{\ll}
\roundedpicture[width = \ww, height = \hh]{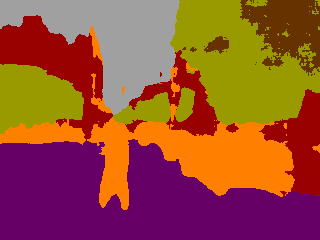}&\hspace{\ll}
\roundedpicture[width = \ww, height = \hh]{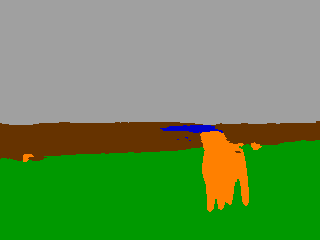}&\hspace{\ll}
\roundedpicture[width = \ww, height = \hh]{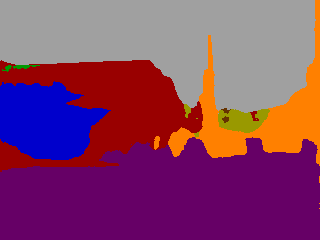}\\

\roundedpicture[width = \ww, height = \hh]{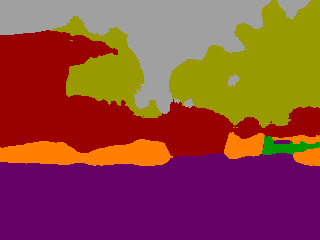}&\hspace{\ll}
\roundedpicture[width = \ww, height = \hh]{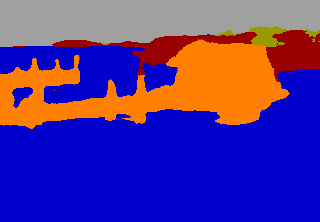}&\hspace{\ll}
\roundedpicture[width = \ww, height = \hh]{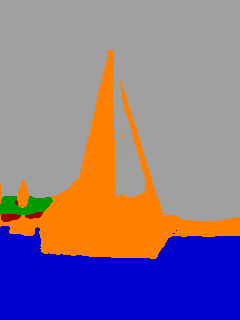}&\hspace{\ll}
\roundedpicture[width = \ww, height = \hh]{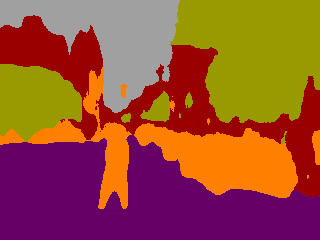}&\hspace{\ll}
\roundedpicture[width = \ww, height = \hh]{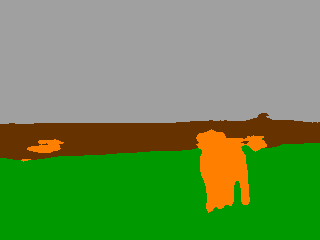}&\hspace{\ll}
\roundedpicture[width = \ww, height = \hh]{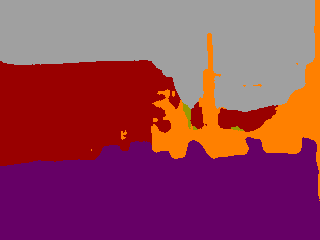}\\

\roundedpicture[width = \ww, height = \hh]{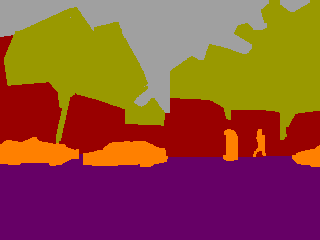}&\hspace{\ll}
\roundedpicture[width = \ww, height = \hh]{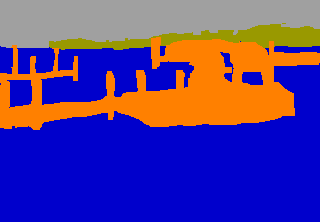}&\hspace{\ll}
\roundedpicture[width = \ww, height = \hh]{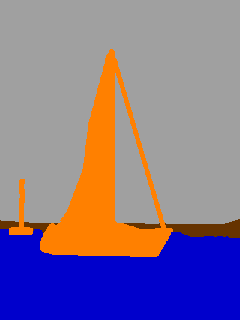}&\hspace{\ll}
\roundedpicture[width = \ww, height = \hh]{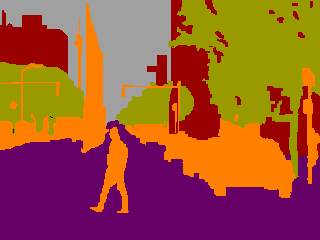}&\hspace{\ll}
\roundedpicture[width = \ww, height = \hh]{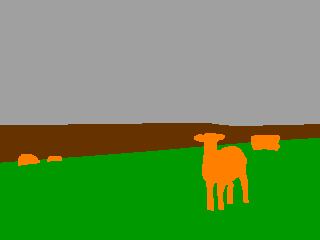}&\hspace{\ll}
\roundedpicture[width = \ww, height = \hh]{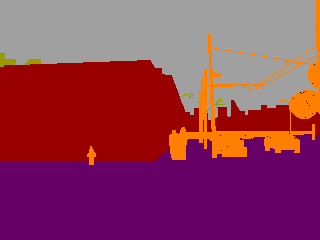}\\
\end{tabular}
\def \ww{0.035\textwidth}
\def \hh{0.015\textwidth}
\begin{tabular}{ccccccccccccccc}
\small Legend:&
\includegraphics[width=\ww, height = \hh]{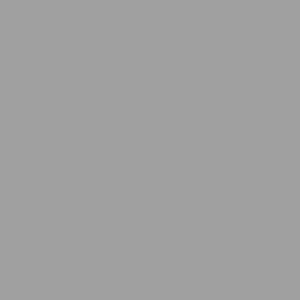}&\hspace{\ll} \small Sky&
\includegraphics[width=\ww, height = \hh]{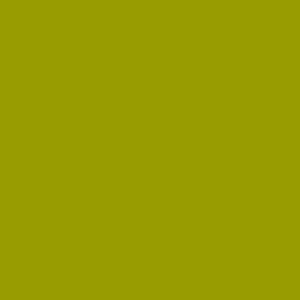}&\hspace{\ll} \small Tree&
\includegraphics[width=\ww, height = \hh]{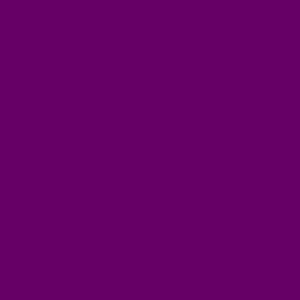}&\hspace{\ll} \small Road&
\includegraphics[width=\ww, height = \hh]{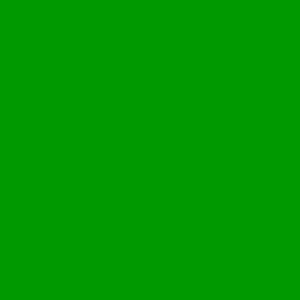}&\hspace{\ll} \small Grass&
\includegraphics[width=\ww, height = \hh]{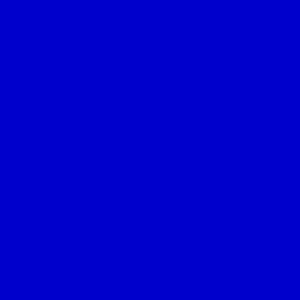}&\hspace{\ll} \small Water&
\includegraphics[width=\ww, height = \hh]{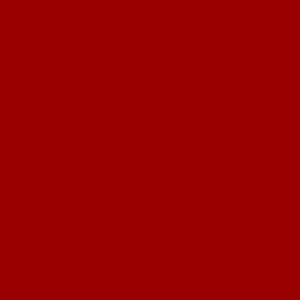}&\hspace{\ll} \small Building&
\includegraphics[width=\ww, height = \hh]{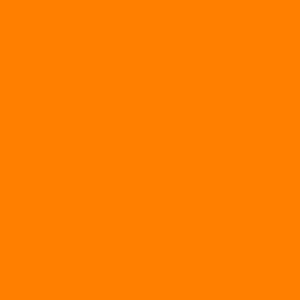}&\hspace{\ll} \small Foreground\\
\end{tabular}
\caption{Test samples of scene labeling on Stanford background dataset~\cite{Gould2009}. \textbf{First row:} Input image, \textbf{second row:} CHM, \textbf{third row:} CHM with intra-class connection, \textbf{Fourth row:} Groundtruth. Using intra-class contextual information improves the performance.}
\label{Fig:SBDsamples}
\end{figure*}

We trained eight CHMs in a one-vs-all architecture.
To take advantage of intra-class contextual information, we allowed CHMs to communicate with each other at three upper levels of the hierarchy.
At those levels, classifiers get samples of context images of other classes as well as their own class.
The performance of CHM with and without intra-class connection is reported in Table~\ref{Tab:SBD}.
\begin{table}
\caption{Testing performance of different methods on Stanford background dataset~\cite{Gould2009}: Pixelwise accuracy, class-average accuracy, and computation time. }
\centering
\begin{tabular}{cccc} \toprule
Method & Pixel Acc. & Class Acc. & CT (sec.)  \\ \midrule
Region-based Energy~\cite{Gould2009}&$76.4\%$&$?$&$10-600$\\ \midrule
Selecting Regions~\cite{Kumar2010Efficiently}&$79.4\%$&$?$&$600$\\ \midrule
\shortstack{Stacked Hierarchical\\ Labeling~\cite{Munoz2010Stacked}}&$76.9\%$&$66.2\%$&$12$\\ \midrule
Superparsing~\cite{Tighe2010Superparsing}&$77.5\%$&$?$&$10$\\ \midrule
\shortstack{Recursive Neural\\ Networks~\cite{Socher2011Parsing}}&$78.1\%$&$?$&$?$\\ \midrule
Pylon Model~\cite{Lempitsky2011Pylon}&$81.9\%$&$72.4\%$&$60$\\ \midrule
Ren \textit{et al.}~\cite{RenRGB2012}&$\bf{82.9\%}$&$74.5\%$&$?$\\ \midrule
Singlescale ConvNet~\cite{Farabet2013}&$66\%$&$56.5\%$&$\bf{0.35}$\\ \midrule
Multiscale ConvNet~\cite{Farabet2013}&$78.8\%$&$72.4\%$&$0.6$\\ \midrule
\shortstack{Multiscale ConvNet+\\CRF on gPb~\cite{Farabet2013}}&$81.4\%$&$\bf{76.0\%}$&$60.5$\\ \midrule
CHM&$82.30\%$&$73.70\%$&$60$\\ \midrule
\shortstack{CHM with Intra-class \\Connection}&$\bf{82.95\%}$&$74.32\%$&$65$\\ \bottomrule
\end{tabular}
\label{Tab:SBD}
\end{table}
Our CHM achieves state-of-the-art performance in terms of pixel accuracy.
Due to the absence of any global constraint for label consistency, CHM performs worse than~\cite{RenRGB2012,Farabet2013} in terms of class-average accuracy.
Similar to~\cite{Farabet2013}, we computed superpixels~\cite{Felzenszwalb2004Efficient} for each image and then assign the most common label, based on CHM output, to each superpixel.
Unlike~\cite{Farabet2013}, this approach had negligible impact on the performance and improved the pixel accuracy only to $83\%$.
This shows CHM is a powerful pixel classifier.
In our experiment, inference took about $65$ seconds for each image (half of it was spent on computing the features).
A few test samples of the Stanford background dataset and corresponding CHM results are shown in Figure~\ref{Fig:SBDsamples}.
Using intra-class connection improves the label consistency in the results.

\begin{figure}
\centering
\includegraphics[width = .9\linewidth]{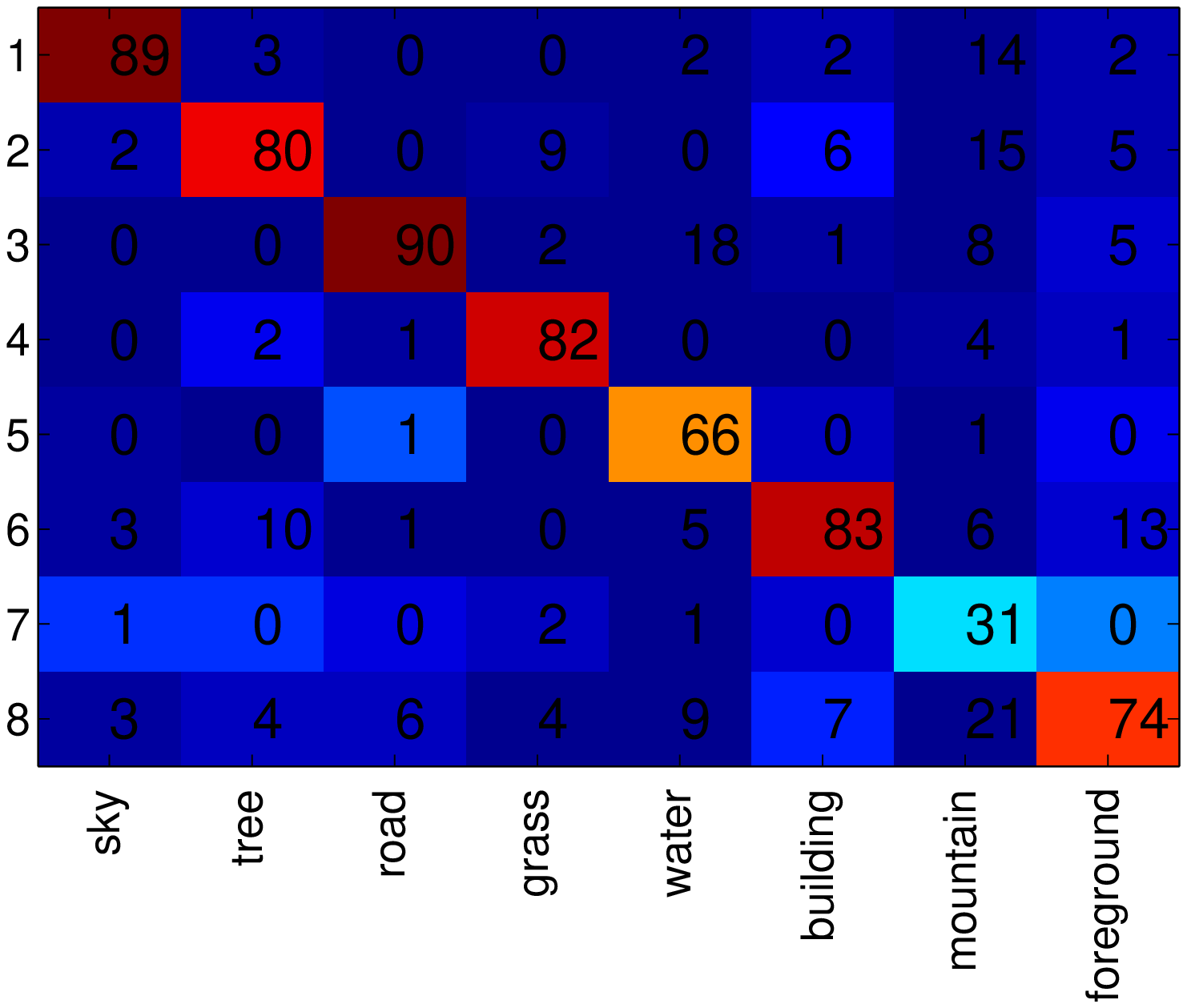}
\caption{The confusion matrix of CHM results on the Stanford background dataset~\cite{Gould2009}. The overall class-average accuracy is $74.32\%$.}
\label{Fig:conf}
\end{figure}

The $8$-class confusion matrix of CHM is shown in Figure~\ref{Fig:conf}.
The hard classes are mountain, water, and foreground.
This is consistent with the reported results in~\cite{Munoz2010Stacked,RenRGB2012}.
Even though the performance of CHM is similar to~\cite{RenRGB2012} for most of the classes, it performs significantly better on the foreground category compared to~\cite{RenRGB2012} achieving $74.1\%$ vs $63\%$.
\subsection{Edge Detection}
In this section we show the performance of CHM on two edge detection datasets: BSDS 500~\cite{Martin2001} and NYU Depth (v2)~\cite{Silberman2011}.
We used the popular evaluation framework available in the gPb package~\cite{Arbelaez2011Contour} to compare CHM performance with other methods.
The evaluation framework computes three metrics: Fvalue computed with a fixed threshold for the entire dataset (ODS), F-value computed with per-image best thresholds (OIS), and the average precision (AP).

We trained a CHM with $5$ levels for both datasets.
Similar to~\cite{Lim2013Sketch,Dollar2013}, we adopted a multi-scale strategy to compute edge maps.
That is, at test time, we ran the trained CHM on the original, as well as double and half resolution versions of each input image.
We then resized the results to the original image resolution and averaged them to obtain the edge map.
We also used the standard non-maximal suppression, suggested in~\cite{Arbelaez2011Contour,Ren2012,Lim2013Sketch,Dollar2013}, to obtain thinned edges.

\subsubsection{BSDS 500 dataset}
\begin{figure}
\includegraphics[width=0.45\textwidth]{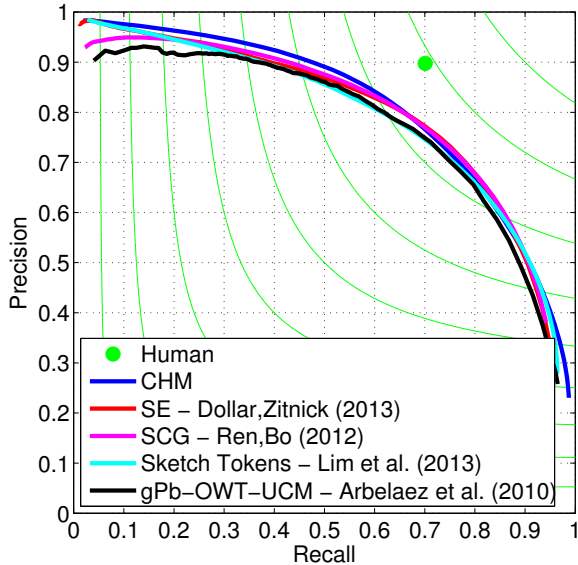}
\caption{Precision-recall curves of CHM in comparison with other methods for BSDS 500 dataset~\cite{Martin2001}.}
\label{Fig:BSDS500}
\end{figure}

\begin{figure*}
\def \ll{-.45cm}
\begin{tabular}{ccccccc}
\roundedpicture[width = .132\textwidth]{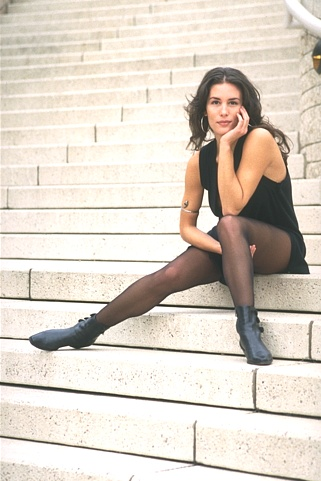}& \hspace{\ll}
\roundedpicture[width = .132\textwidth]{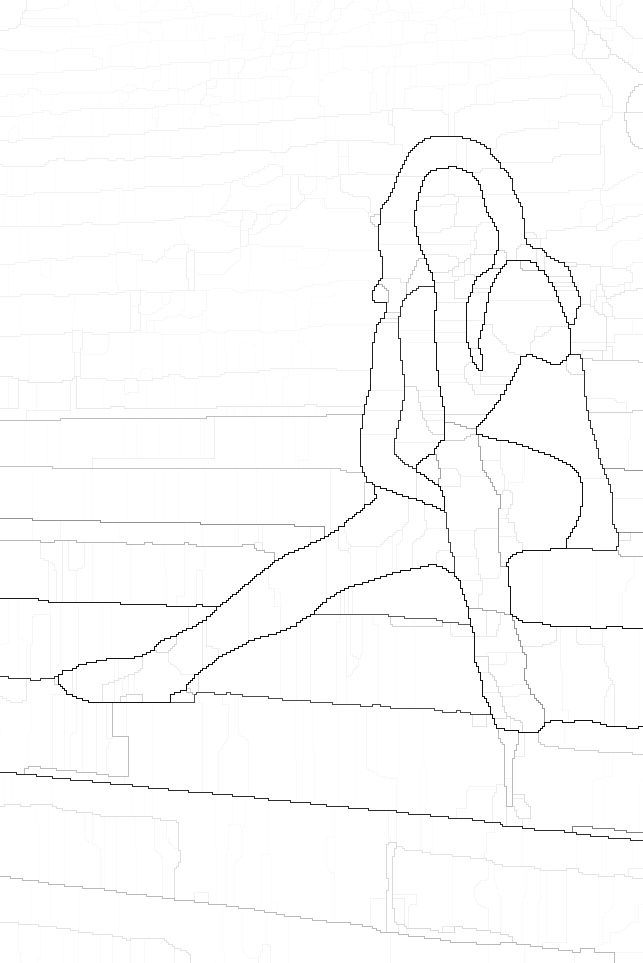}&\hspace{\ll}
\roundedpicture[width = .132\textwidth]{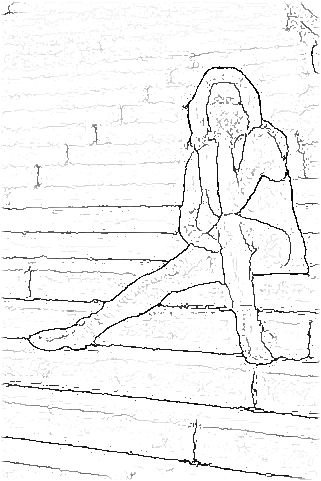}&\hspace{\ll}
\roundedpicture[width = .132\textwidth]{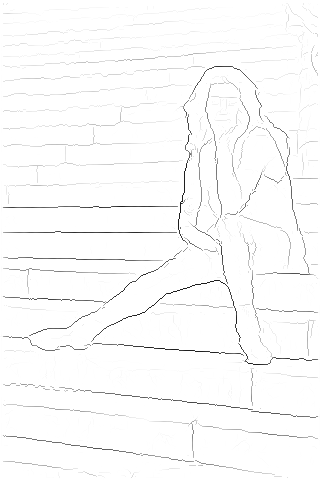}&\hspace{\ll}
\roundedpicture[width = .132\textwidth]{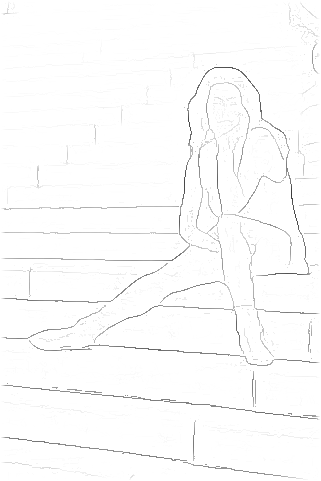}&\hspace{\ll}
\roundedpicture[width = .132\textwidth]{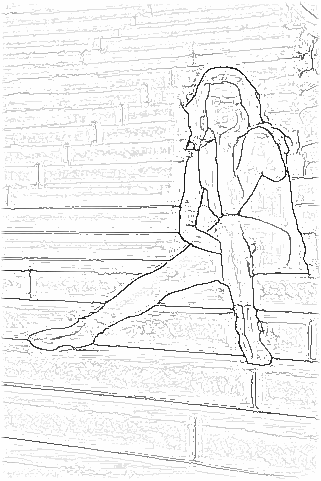}&\hspace{\ll}
\roundedpicture[width = .132\textwidth]{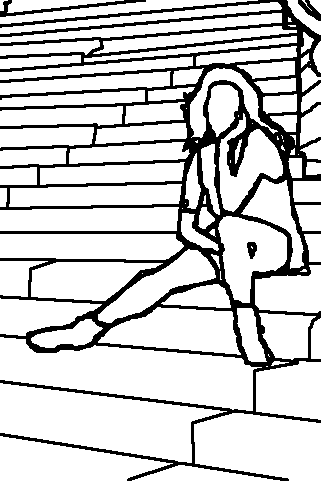}\\ [-.5ex]
\roundedpicture[width = .132\textwidth]{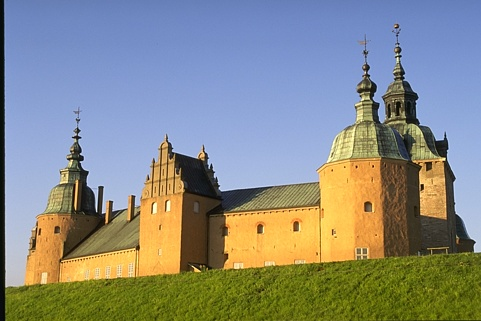}& \hspace{\ll}
\roundedpicture[width = .132\textwidth]{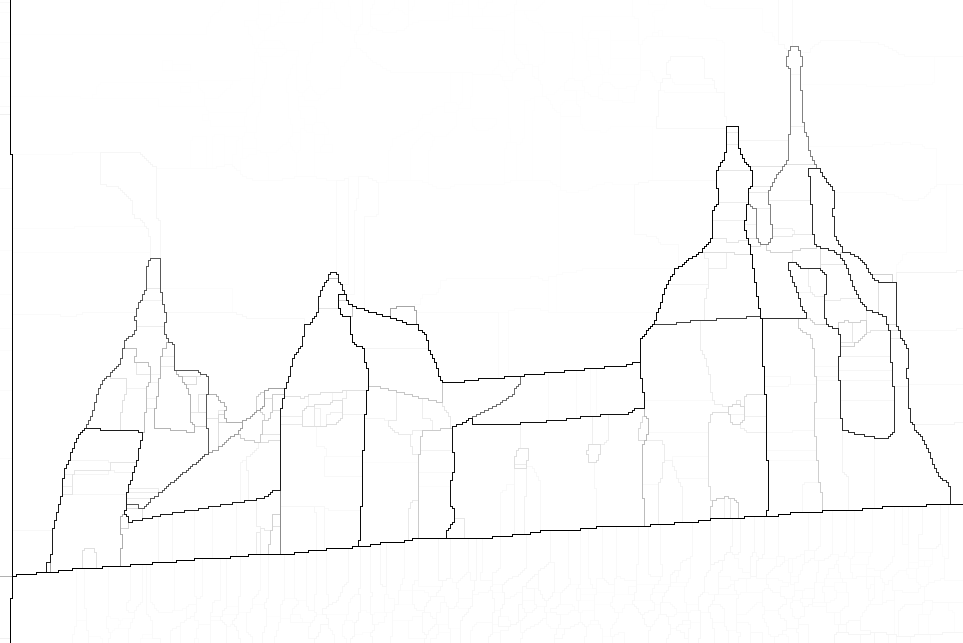}&\hspace{\ll}
\roundedpicture[width = .132\textwidth]{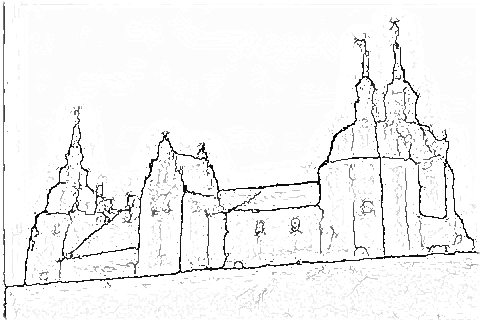}&\hspace{\ll}
\roundedpicture[width = .132\textwidth]{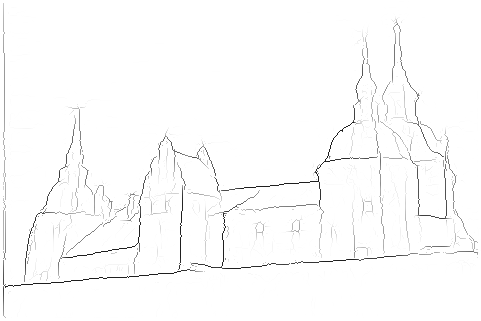}&\hspace{\ll}
\roundedpicture[width = .132\textwidth]{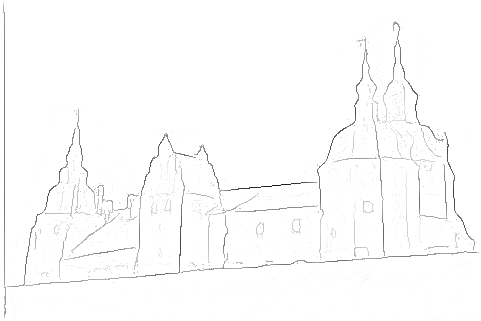}&\hspace{\ll}
\roundedpicture[width = .132\textwidth]{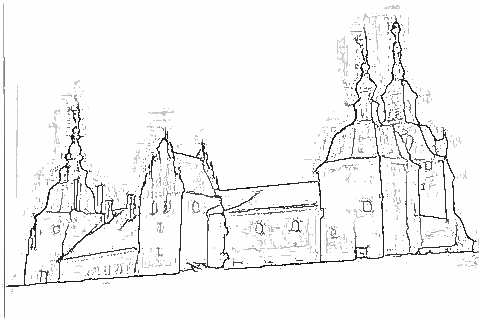}&\hspace{\ll}
\roundedpicture[width = .132\textwidth]{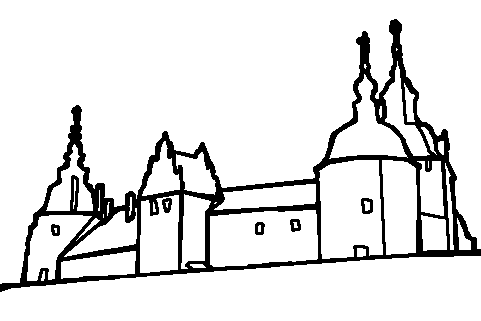}\\[-.5ex]
\roundedpicture[width = .132\textwidth]{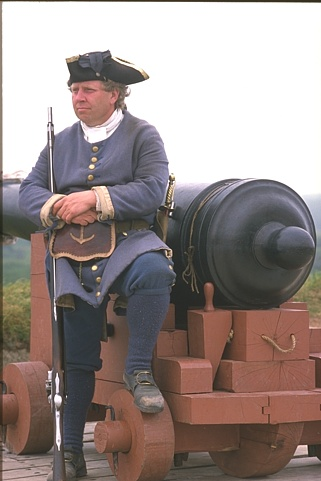}& \hspace{\ll}
\roundedpicture[width = .132\textwidth]{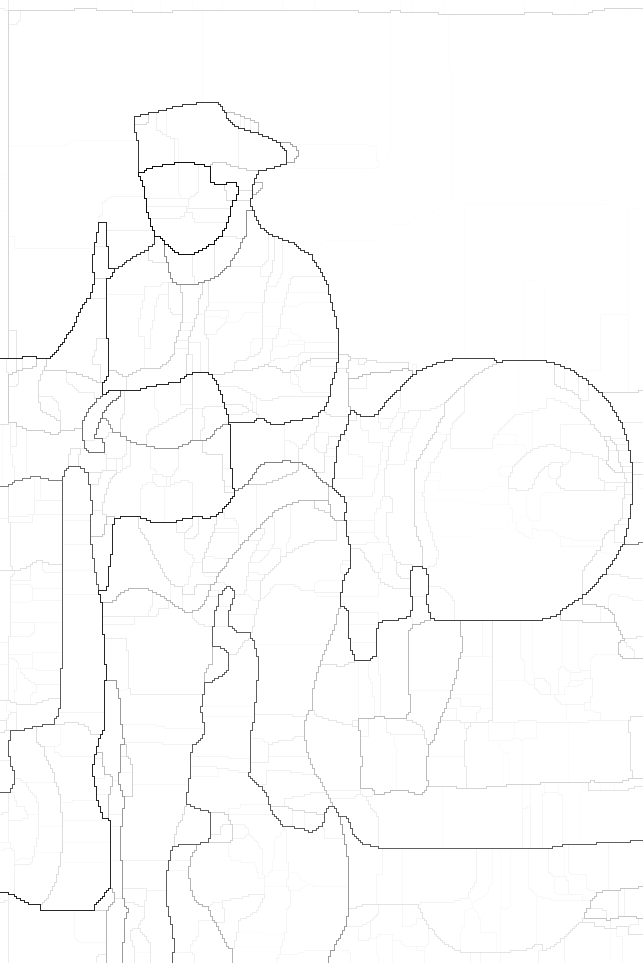}&\hspace{\ll}
\roundedpicture[width = .132\textwidth]{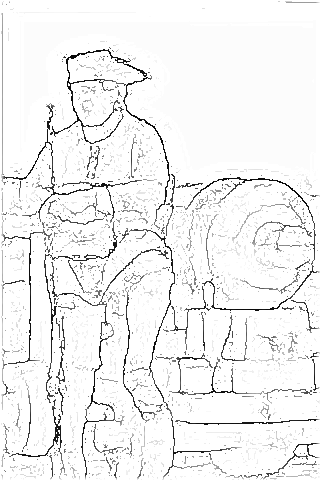}&\hspace{\ll}
\roundedpicture[width = .132\textwidth]{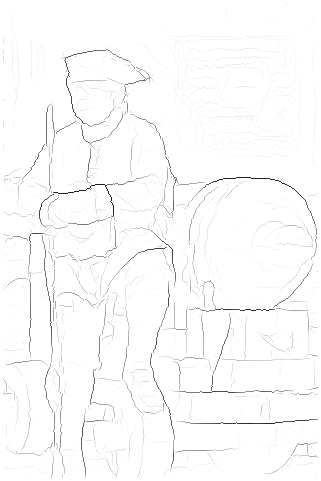}&\hspace{\ll}
\roundedpicture[width = .132\textwidth]{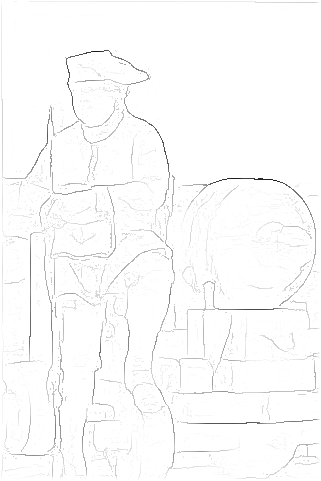}&\hspace{\ll}
\roundedpicture[width = .132\textwidth]{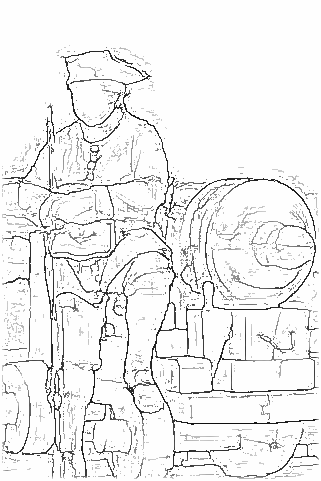}&\hspace{\ll}
\roundedpicture[width = .132\textwidth]{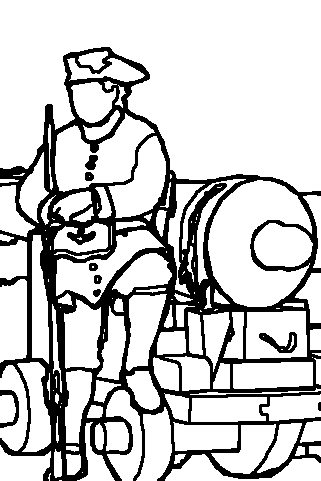}\\[-.5ex]
\roundedpicture[width = .132\textwidth]{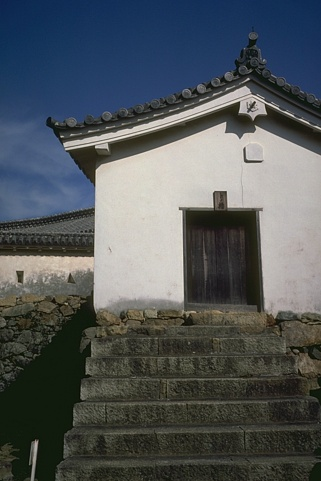}& \hspace{\ll}
\roundedpicture[width = .132\textwidth]{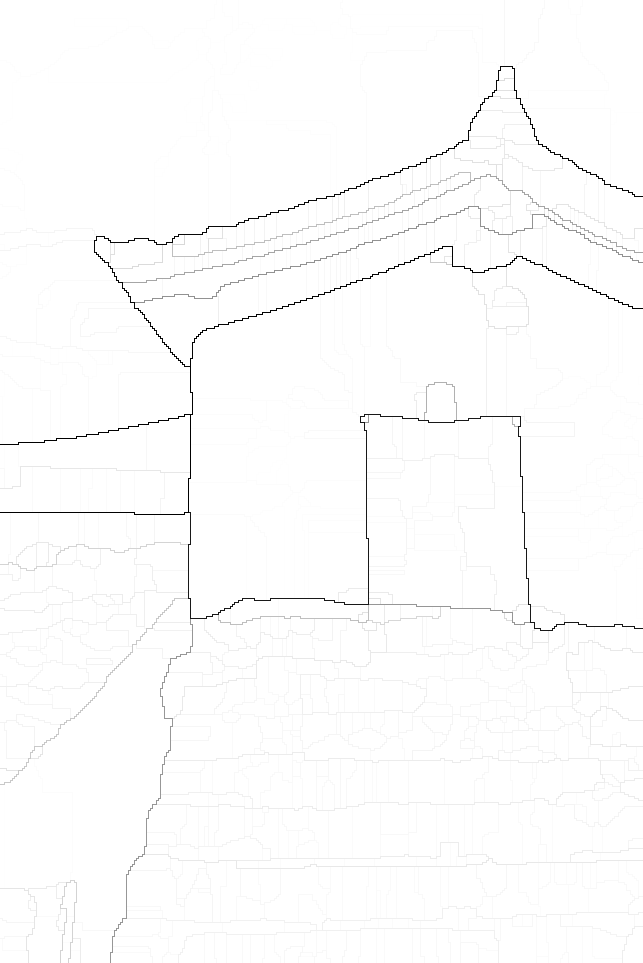}&\hspace{\ll}
\roundedpicture[width = .132\textwidth]{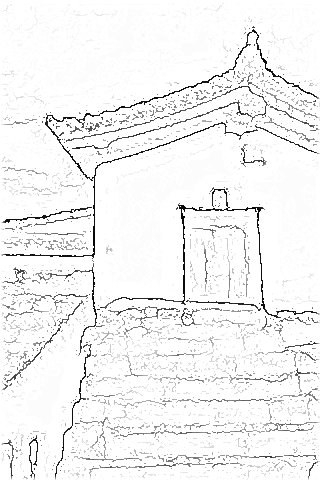}&\hspace{\ll}
\roundedpicture[width = .132\textwidth]{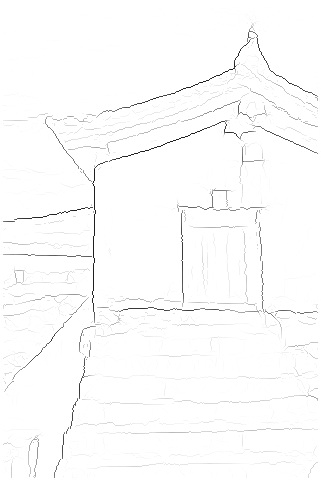}&\hspace{\ll}
\roundedpicture[width = .132\textwidth]{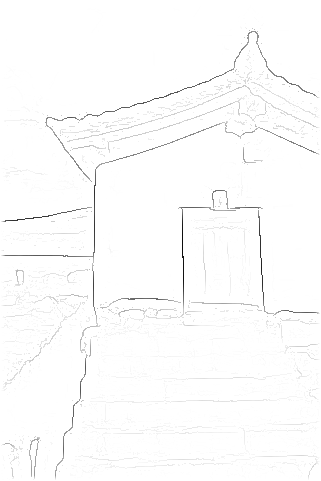}&\hspace{\ll}
\roundedpicture[width = .132\textwidth]{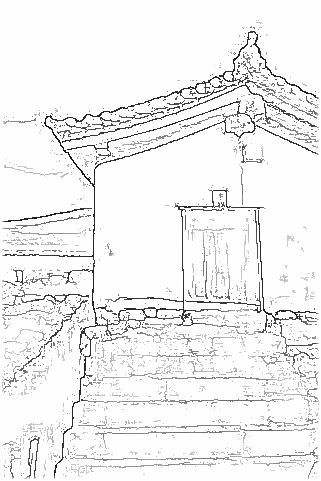}&\hspace{\ll}
\roundedpicture[width = .132\textwidth]{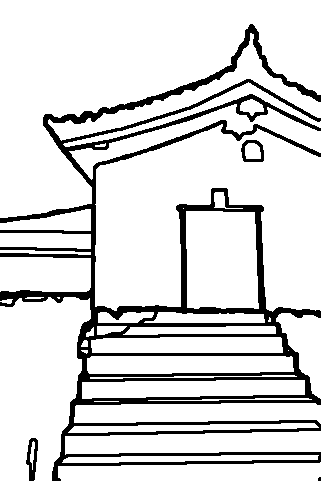}\\[-.5ex]
\roundedpicture[width = .132\textwidth]{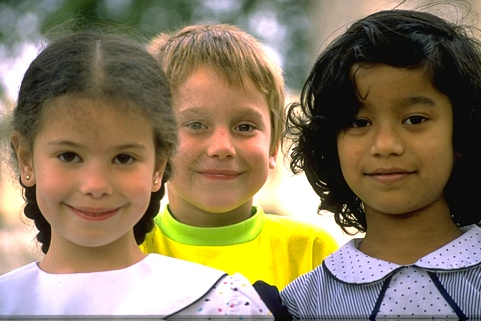}& \hspace{\ll}
\roundedpicture[width = .132\textwidth]{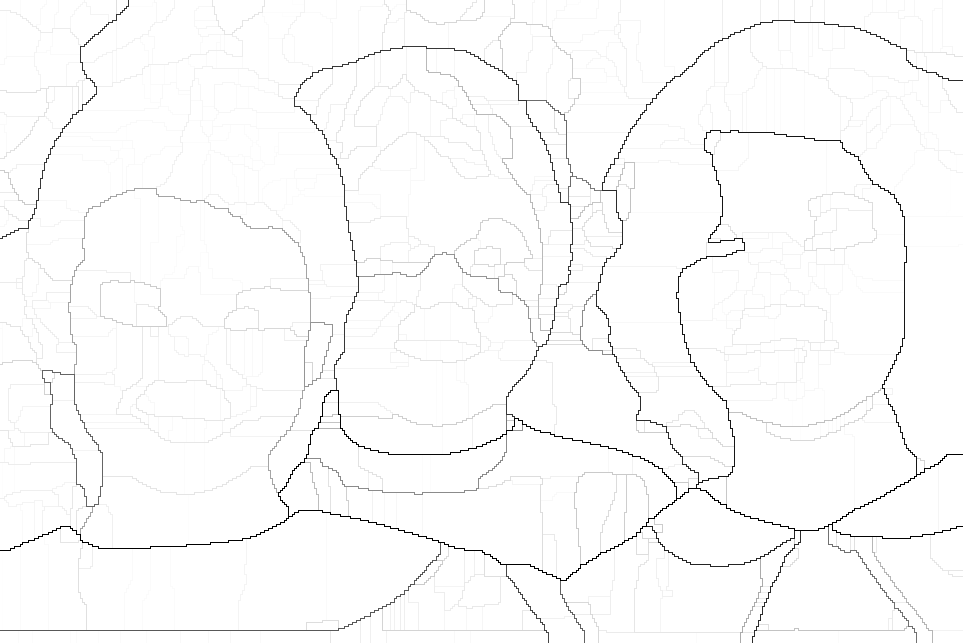}&\hspace{\ll}
\roundedpicture[width = .132\textwidth]{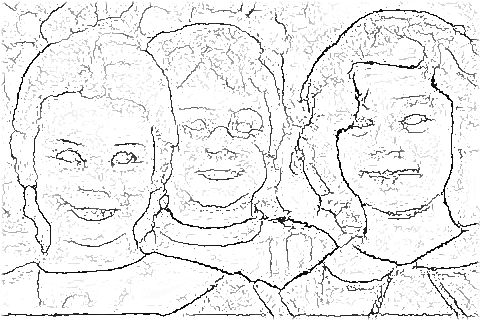}&\hspace{\ll}
\roundedpicture[width = .132\textwidth]{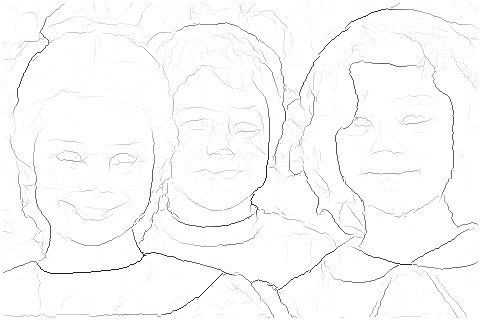}&\hspace{\ll}
\roundedpicture[width = .132\textwidth]{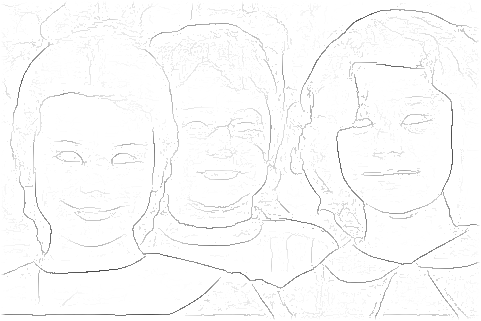}&\hspace{\ll}
\roundedpicture[width = .132\textwidth]{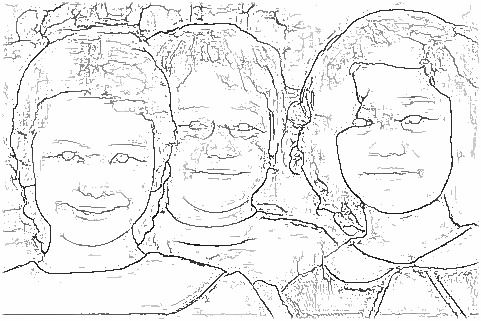}&\hspace{\ll}
\roundedpicture[width = .132\textwidth]{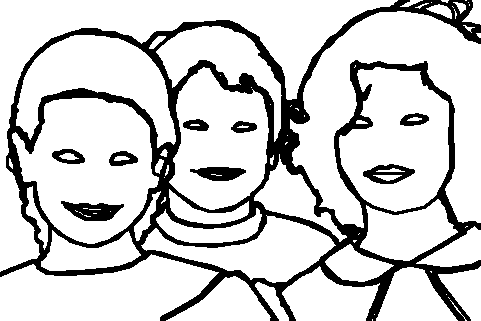}\\[-.5ex]
(a)&(b) &(c)&(d)&(e) &(f) &(g)\\
\end{tabular}
\caption{Test samples of edge detection on BSDS 500~\cite{Martin2001} dataset. (a) Input image, (b) gPb-OWT-UCM~\cite{Arbelaez2011Contour}, (c) Sketch tokens~\cite{Lim2013Sketch}, (d) SCG~\cite{Ren2012}, (e) SE~\cite{Dollar2013}, (f) CHM, (g) Groundtruth. CHM is able to capture finer details like upper stairs in the first row, steeples in the second row, and wheels in the third row.}
\label{Fig:BSDSsamples}
\end{figure*}

Berkeley segmentation dataset and benchmarks (BSDS 500)~\cite{Martin2001,Arbelaez2011Contour} is an extension of BSDS 300 dataset and used widely for the evaluation of edge detection techniques.
It contains $200$ training, $100$ validation, and $200$ testing images of resolution $321 \times 481$ pixels (roughly).
The human annotations for each image is included in the dataset.
The precision-recall curves for CHM and four other methods are shown in Figure~\ref{Fig:BSDS500}.
Note that CHM achieves high precision and recall at both ends of the precision-recall curve.
The evaluation metrics are reported in Table~\ref{Tab:BSDS500}.
\begin{table}
\caption{Testing performance of different methods on BSDS 500 dataset~\cite{Martin2001}. CHM achieves near state-of-the-art performance in terms of ODS and OIS, and improves over other methods significantly in terms of AP. }
\centering
\begin{tabular}{cccc} \toprule
Method & ODS & OIS & AP  \\ \midrule
gPb-OWT-UCM~\cite{Arbelaez2011Contour}&$0.726$&$0.760$&$0.727$\\ \midrule
Sketch Tokens~\cite{Lim2013Sketch}&$0.728$&$0.746$&$0.780$\\ \midrule
SCG~\cite{Ren2012}&$0.739$&$0.758$&$0.773$\\ \midrule
SE~\cite{Dollar2013}&$\bf{0.741}$&$\bf{0.760}$&$0.780$\\ \midrule
CHM&$0.735$&$0.751$&$\bf{0.804}$\\ \bottomrule
\end{tabular}
\label{Tab:BSDS500}
\end{table}
While CHM performs about the same as SCG~\cite{Ren2012} and SE~\cite{Dollar2013} in terms od ODS and OIS, it achieves state-of-the-art performance in terms of AP.
It must be emphasized that unlike gPb~\cite{Arbelaez2011Contour} and SCG~\cite{Ren2012}, our CHM does not include any globalization step and only relies on the local patch information.
In addition, our CHM is a general patch-based model and unlike gPb~\cite{Arbelaez2011Contour}, SCG~\cite{Ren2012}, and SE~\cite{Dollar2013} can be used in general scene labeling frameworks.
Finally we will show in section~\ref{Sec:generalization} that the cross-dataset generalization performance of CHM is significantly better than other learning-based approaches, \textit{i.e.,} sketch tokens~\cite{Lim2013Sketch}, SCG~\cite{Ren2012}, and SE~\cite{Dollar2013}.
A few test examples of BSDS 500 dataset and corresponding edge detection results are shown in Figure~\ref{Fig:BSDSsamples}. As shown in our results, CHM captures finer details such as upper stairs in the first row, steeples in the second row, and wheels in the third row.

\subsubsection{NYU depth dataset (v2)}
\begin{figure}
\includegraphics[width=0.45\textwidth]{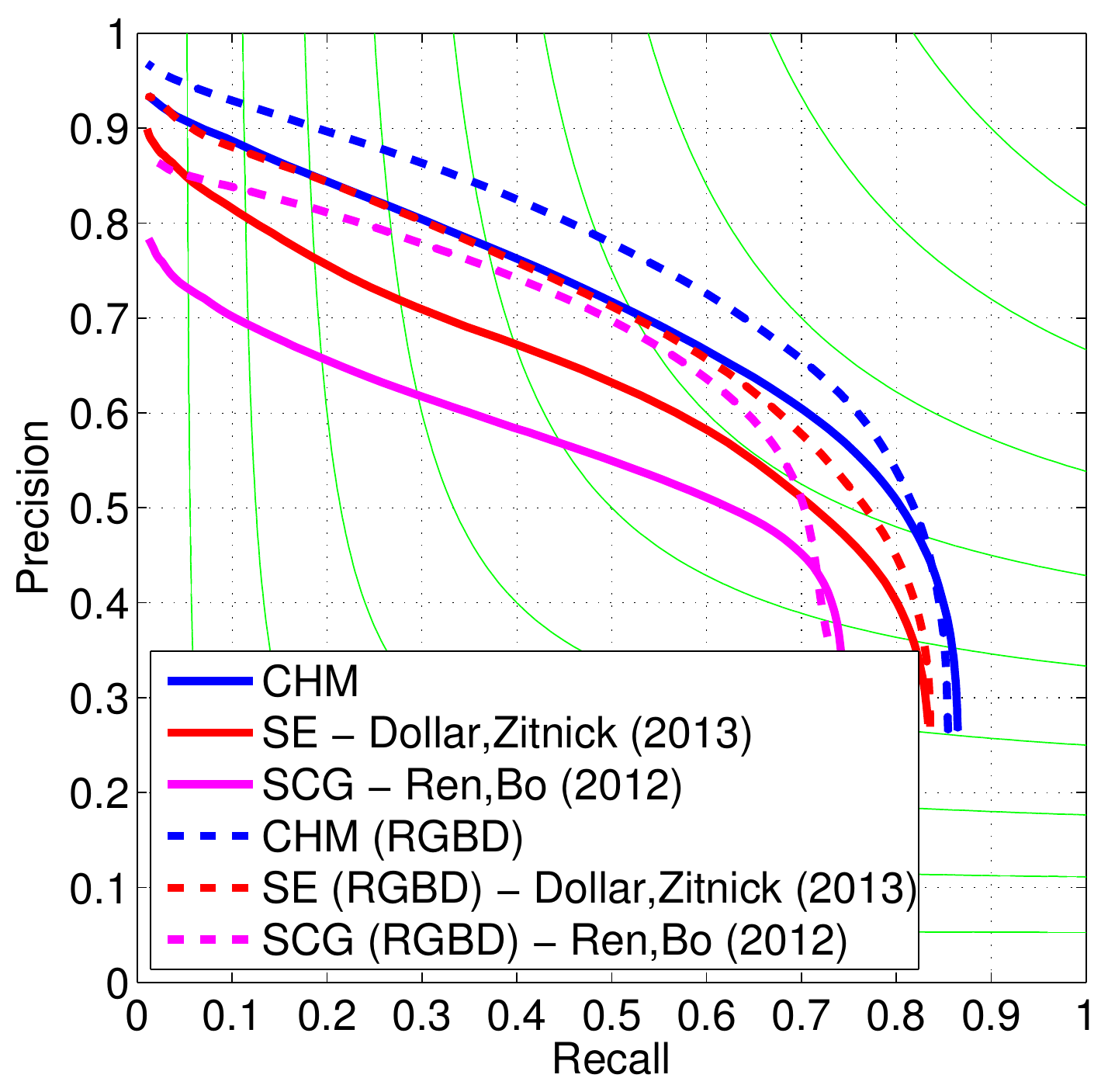}
\caption{Precision-recall curves of different methods for NYU depth dataset~\cite{Silberman2011} using RGB (solid lines) and RGBD(dashed lines) modalities.}
\label{Fig:NYU}
\end{figure}

\begin{figure*}
\def \ll{-.48cm}
\def \ww{.119\textwidth}
\begin{tabular}{cccccccc}
\roundedpicture[width = \ww]{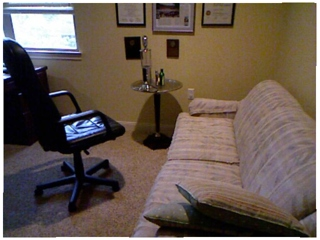}& \hspace{\ll}
\roundedpicture[width = \ww]{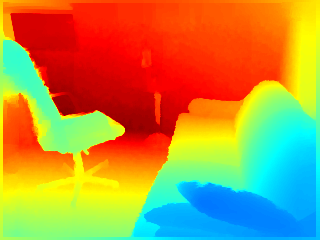}& \hspace{\ll}
\roundedpicture[width = \ww]{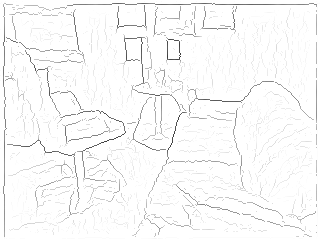}& \hspace{\ll}
\roundedpicture[width = \ww]{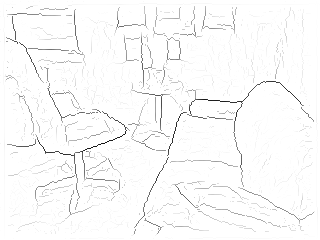}& \hspace{\ll}
\roundedpicture[width = \ww]{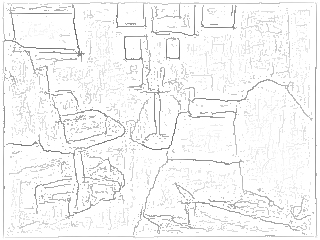}& \hspace{\ll}
\roundedpicture[width = \ww]{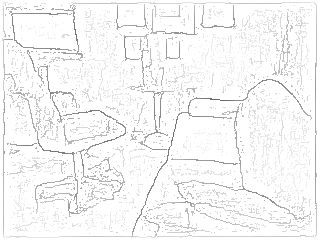}& \hspace{\ll}
\roundedpicture[width = \ww]{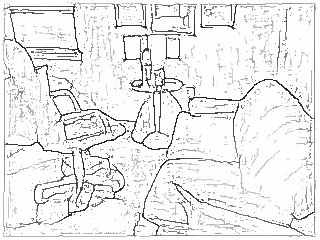}& \hspace{\ll}
\roundedpicture[width = \ww]{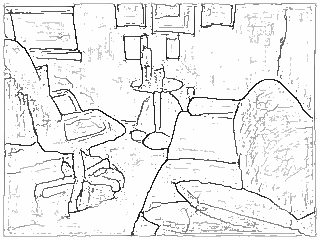}\\[-.5ex]
\roundedpicture[width = \ww]{}& \hspace{\ll}
\roundedpicture[width = \ww]{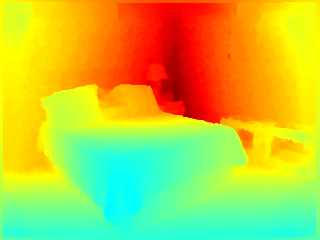}& \hspace{\ll}
\roundedpicture[width = \ww]{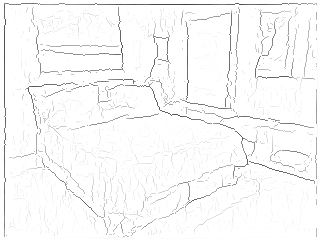}& \hspace{\ll}
\roundedpicture[width = \ww]{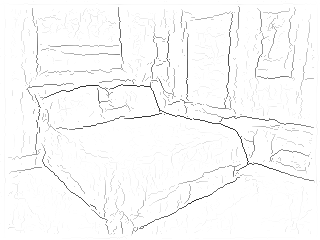}& \hspace{\ll}
\roundedpicture[width = \ww]{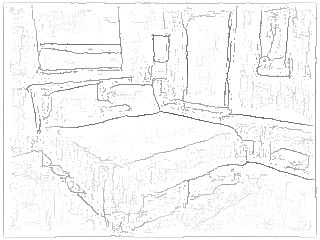}& \hspace{\ll}
\roundedpicture[width = \ww]{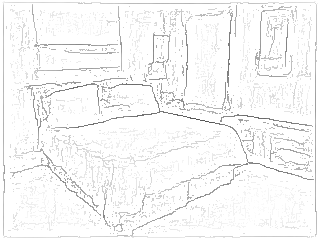}& \hspace{\ll}
\roundedpicture[width = \ww]{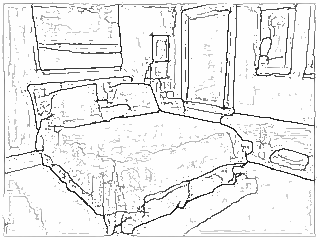}& \hspace{\ll}
\roundedpicture[width = \ww]{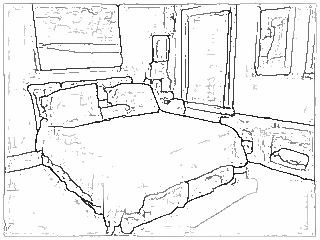}\\[-.5ex]
\roundedpicture[width = \ww]{}& \hspace{\ll}
\roundedpicture[width = \ww]{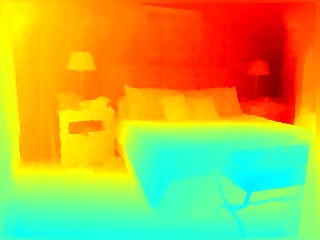}& \hspace{\ll}
\roundedpicture[width = \ww]{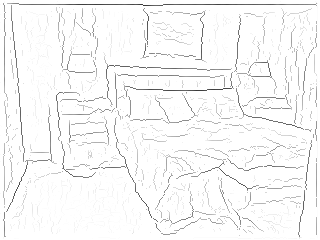}& \hspace{\ll}
\roundedpicture[width = \ww]{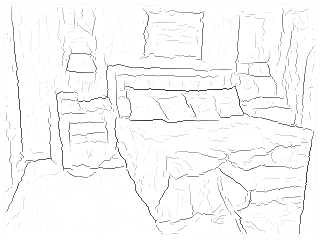}& \hspace{\ll}
\roundedpicture[width = \ww]{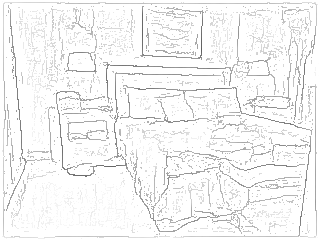}& \hspace{\ll}
\roundedpicture[width = \ww]{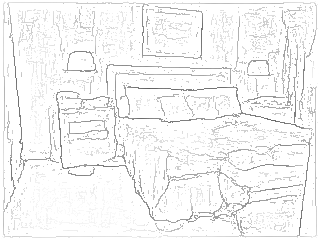}& \hspace{\ll}
\roundedpicture[width = \ww]{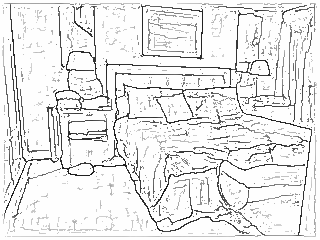}& \hspace{\ll}
\roundedpicture[width = \ww]{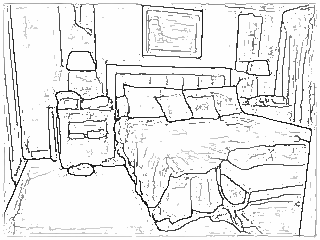}\\[-.5ex]
\roundedpicture[width = \ww]{}& \hspace{\ll}
\roundedpicture[width = \ww]{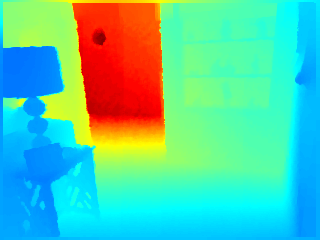}& \hspace{\ll}
\roundedpicture[width = \ww]{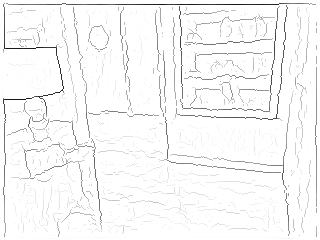}& \hspace{\ll}
\roundedpicture[width = \ww]{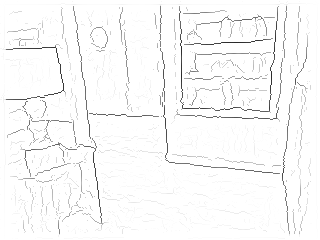}& \hspace{\ll}
\roundedpicture[width = \ww]{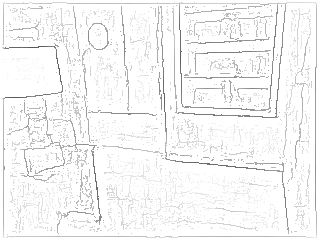}& \hspace{\ll}
\roundedpicture[width = \ww]{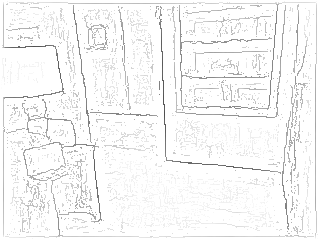}& \hspace{\ll}
\roundedpicture[width = \ww]{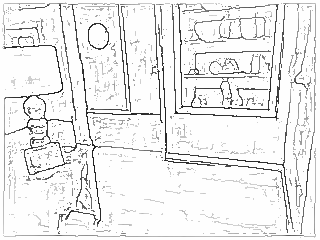}& \hspace{\ll}
\roundedpicture[width = \ww]{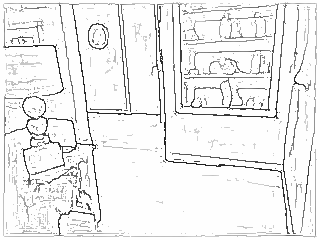}\\[-.5ex]
\roundedpicture[width = \ww]{}& \hspace{\ll}
\roundedpicture[width = \ww]{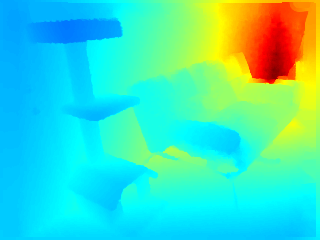}& \hspace{\ll}
\roundedpicture[width = \ww]{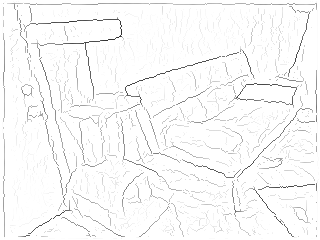}& \hspace{\ll}
\roundedpicture[width = \ww]{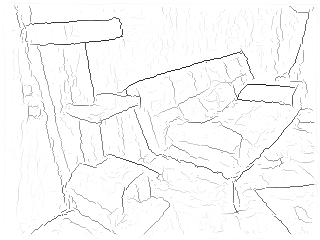}& \hspace{\ll}
\roundedpicture[width = \ww]{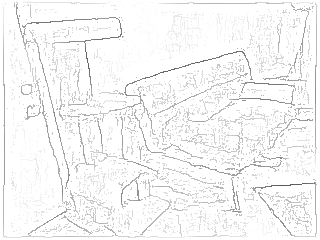}& \hspace{\ll}
\roundedpicture[width = \ww]{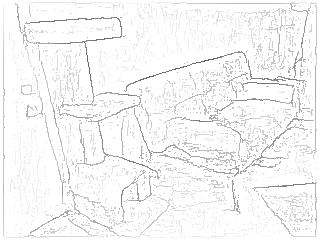}& \hspace{\ll}
\roundedpicture[width = \ww]{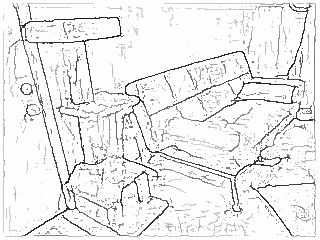}& \hspace{\ll}
\roundedpicture[width = \ww]{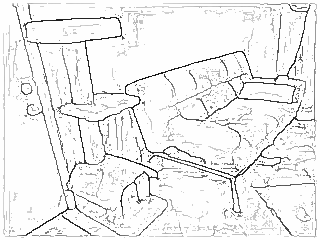}\\[-.5ex]
(a)&(b) &(c)&(d)&(e) &(f) &(g)&(h)\\
\end{tabular}
\caption{Test samples of edge detection on NYU depth (v2) dataset~\cite{Silberman2011}. (a) Input image, (b) Depth image, (c) SCG (RGB)~\cite{Ren2012}, (d) SCG (RGBD)~\cite{Ren2012}, (e) SE (RGB)~\cite{Dollar2013}, (f) SE (RGBD)~\cite{Dollar2013}, (g) CHM (RGB), and (h) CHM (RGBD).}
\label{Fig:NYUsamples}
\end{figure*}

The NYU depth dataset (v2)~\cite{Silberman2011} is an RGB-D dataset containing $1449$ pairs of RGB and depth images of resolution $480 \times 640$ pixels, with corresponding groundtruth semantic segmentations.
We used the scripts provided by the authors of~\cite{Ren2012} to adopt this dataset for edge detection\footnote{The scripts are available at \url{http://homes.cs.washington.edu/~xren/research/nips2012/sparse_contour_gradients_v1.1.zip}}.
They used $60\%$ of the images for training ($869$ images) and the remaining $40\%$ for testing ($580$ images).
The images were also resized to $240\times 320$ resolution.
We evaluated the performance of CHM using RGB and RGBD modalities.
For the depth channel, we computed the same set of features that we extract from the RGB color channels.
In Table~\ref{Tab:NYU}, we compare CHM with SCG~\cite{Ren2012} and SE~\cite{Dollar2013}.
\begin{table}
\caption{Testing performance of different methods on NYU depth dataset~\cite{Silberman2011} using RGB (top), and RGBD (bottom) modalities. CHM achieves state-of-the-art performance for both cases.}
\centering
\begin{tabular}{cccc} \toprule
Method & ODS & OIS & AP  \\ \midrule
SCG~\cite{Ren2012} (RGB)&$0.557$&$0.569$&$0.438$\\ \midrule
SE~\cite{Dollar2013} (RGB) &$0.596$&$0.608$&$0.541$\\ \midrule
CHM (RGB)&$\bf{0.649}$&$\bf{0.661}$&$\bf{0.625}$\\ \midrule [.5ex]
SCG~\cite{Ren2012} (RGBD)&$0.621$&$0.632$&$0.534$\\ \midrule
SE~\cite{Dollar2013} (RGBD)&$0.636$&$0.647$&$0.601$\\ \midrule
CHM (RGBD) &$\bf{0.678}$&$\bf{0.690}$&$\bf{0.665}$\\ \bottomrule
\end{tabular}
\label{Tab:NYU}
\end{table}
CHM performs significantly better than other methods and reaches an F-value of $0.649$ for RGB and $0.678$ for RGBD.
The precision-recall curves are shown in Figure~\ref{Fig:NYU} and qualitative comparisons are shown in Figure~\ref{Fig:NYUsamples}.
\subsubsection{Cross-dataset generalization} \label{Sec:generalization}
\begin{figure}
\includegraphics[width=0.45\textwidth]{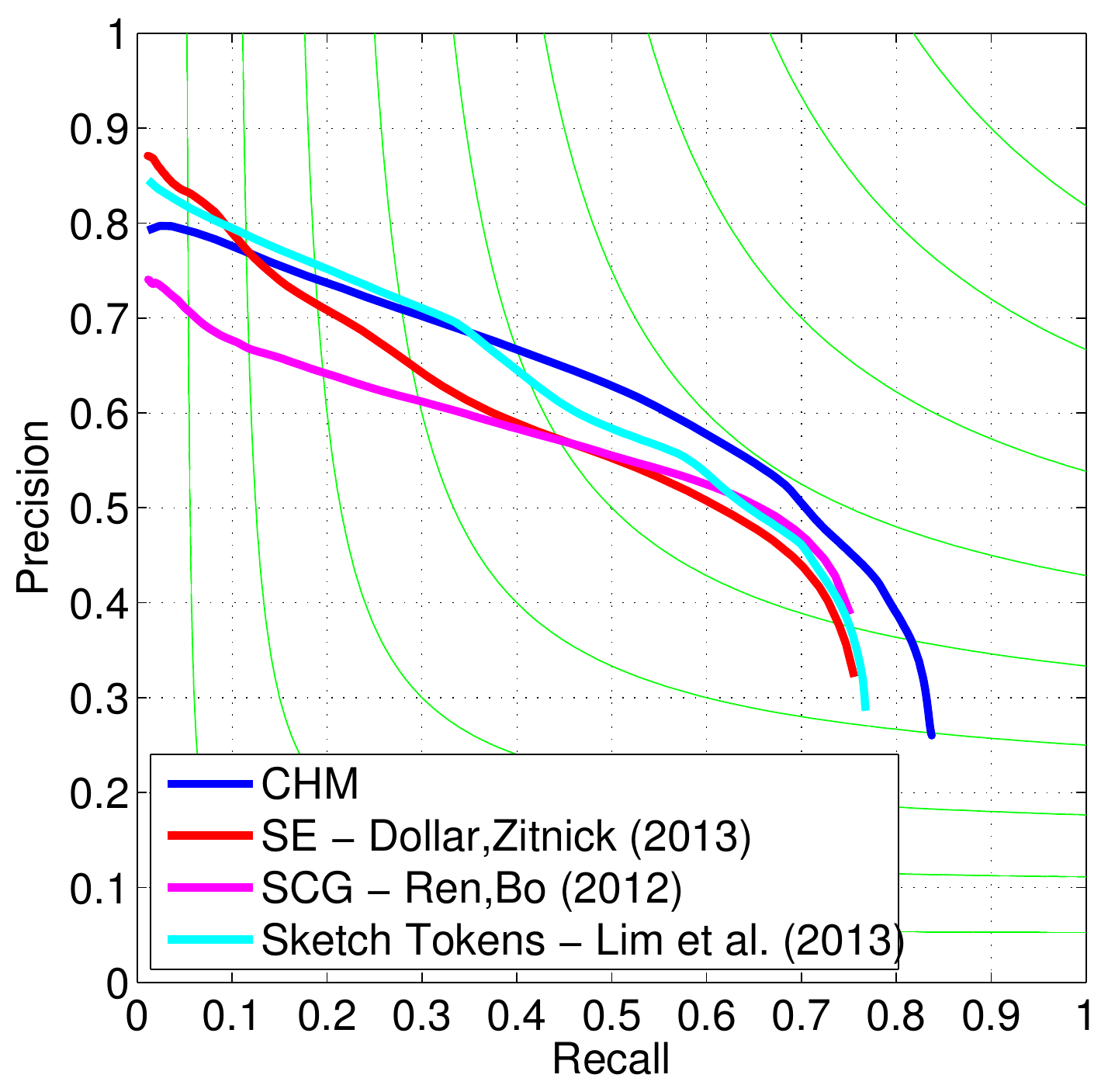}
\caption{Precision-recall curves of different methods for NYU depth dataset~\cite{Silberman2011} using BSDS 500 dataset~\cite{Martin2001} for training. Cross-dataset generalization performance of CHM is better compared to other methods.}
\label{Fig:BSDSNYU}
\end{figure}

Inspired by the work of Dollar and Zitnick~\cite{Dollar2013}, we performed a set of experiments to examine the generalization performance of CHM in comparison to other learning-based methods.
We used the trained CHM on BSDS 500 dataset and ran it on NYU depth dataset for RGB modality.
The authors of sketch tokens~\cite{Lim2013Sketch}, SCG~\cite{Ren2012}, and SE~\cite{Dollar2013} have provided their models for BSDS 500 dataset; so, we could run the same experiment for their methods.
The performance metrics for different methods are reported in Table~\ref{Tab:BSDSNYU} and corresponding precision-recall curves are shown in Figure~\ref{Fig:BSDSNYU}.
\begin{table}
\caption{Testing performance of different methods on NYU depth dataset~\cite{Silberman2011} using BSDS 500 dataset~\cite{Martin2001} for training. CHM outperforms other learning-based approaches significantly.}
\centering
\begin{tabular}{cccc} \toprule
Method & ODS & OIS & AP  \\ \midrule
Sketch Tokens~\cite{Lim2013Sketch}&$0.567$&$0.581$&$0.490$\\ \midrule
SCG~\cite{Ren2012}&$0.568$&$0.579$&$0.441$\\ \midrule
SE~\cite{Dollar2013}&$0.552$&$0.566$&$0.462$\\ \midrule
CHM&$\bf{0.595}$&$\bf{0.606}$&$\bf{0.528}$\\ \bottomrule
\end{tabular}
\label{Tab:BSDSNYU}
\end{table}
CHM performs significantly better than other methods.
Note that all methods perform about the same on BSDS 500 dataset (Table~\ref{Tab:BSDS500}).
We believe this asserts that our CHM can be used as a general edge detection technique.

\subsection{Biomedical Image Segmentation}
In the last set of experiments, we applied CHM to the membrane detection problem in electron microscopy (EM) images.
This is a challenging problem because of the noisy texture, complex intracellular structures, and similar local appearances among different objects~\cite{Jain2007,Lucchi2010}.
In these experiments, we used a CHM with $2$ stages and $5$ levels per stage. A $24 \times 24$ LDNN was used as the classifier.
In addition to the feature set described in section~\ref{Sec:features}, we included Radon-like features (RLF)~\cite{Kumar2010}, which proved to be informative for membrane detection.

\subsection{Mouse neuropil dataset}\label{Sec:mouse}
This dataset is a stack of $70$ images from the mouse neuropil acquired using serial block face scanning electron microscopy (SBFSEM~\cite{Denk2004}).
It has a resolution of $10 \times 10 \times 50$ nm/pixel and each $2D$ image is $700$ by $700$ pixels.
An expert anatomist annotated membranes, \textit{i.e.,} cell boundaries, in these images.
From those $70$ images, $14$ images were randomly selected and used for training and the $56$ remaining images were used for testing.
The task is to detect membranes in each $2D$ section.

Since the task is detecting the boundary of cells, we compared our method with two general boundary detection methods, gPb-OWT-UCM (global probability of boundary followed by the oriented watershed transform and ultrametric
contour maps)~\cite{Arbelaez2009} and boosted edge learning (BEL)~\cite{Dollar2006}. The testing results for different methods are given in Table~\ref{Tab:NCMIR}. The CHM-LDNN outperforms the other methods with a notably large margin.

\begin{table}
\caption{Testing performance of different methods for the mouse neuropil and Drosophila VNC datasets.}
\centering
\begin{tabular}{ccccc} \toprule
&\multicolumn{2}{c}{Mouse neuropil}&\multicolumn{2}{c}{Drosophila VNC}\\ \cmidrule(r){2-3} \cmidrule(r){4-5}
Method & F-value & G-mean& F-value & G-mean \\ \midrule
\shortstack{\small gPb-OWT\\ \small -UCM~\cite{Arbelaez2009}}&$45.68\%$&$64.75\%$&$49.90\%$&$69.57\%$\\ \midrule
\small BEL~\cite{Dollar2006}&$71.68\%$&$84.46\%$&$70.21\%$&$84.20\%$\\ \midrule
\small MSANN~\cite{Seyedhosseini2011}&$81.99\%$&$90.48\%$&$78.89\%$&$88.74\%$\\ \midrule

\small CHM&$\bf{86.00\%}$&$\bf{92.48\%}$&$\bf{80.72\%}$&$\bf{90.02\%}$\\ \bottomrule
\end{tabular}
\label{Tab:NCMIR}
\end{table}

\begin{figure*}
\centering
\def \wii{.133\linewidth}
\begin{tabular}{ccccccc}
\roundedpicture[width = \wii]{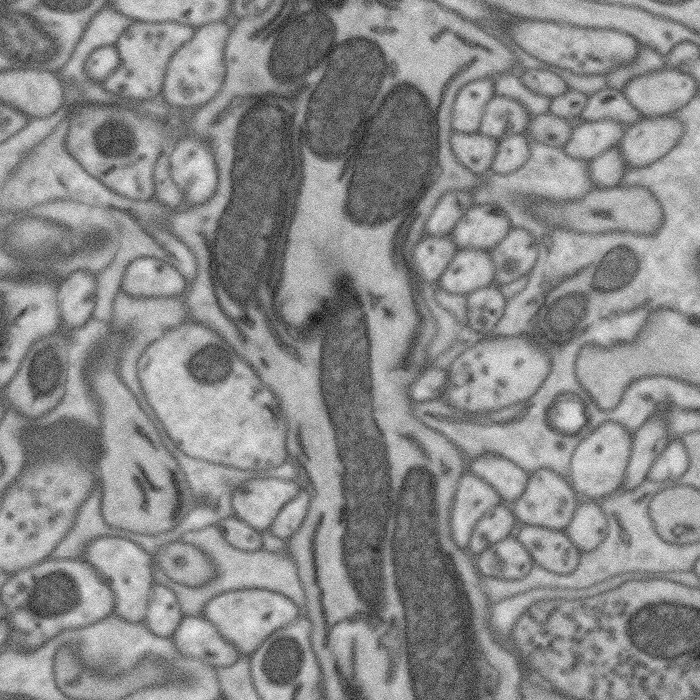}&\hspace{-3.5mm}
\roundedpicture[width = \wii]{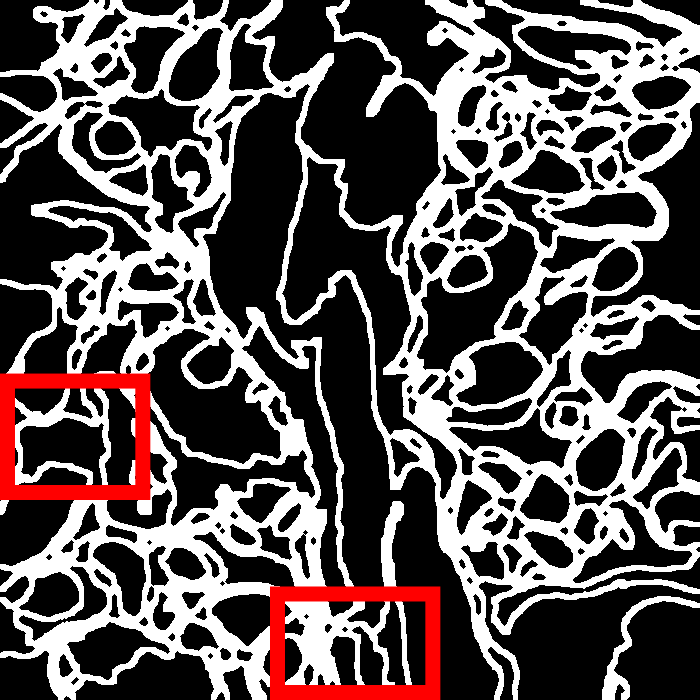} &\hspace{-3.5mm}
\roundedpicture[width = \wii]{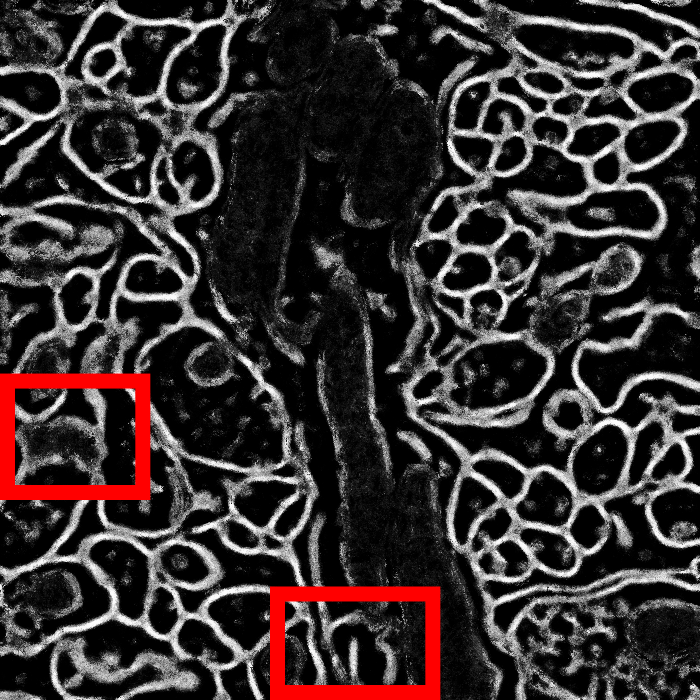}&\hspace{-3.5mm}
\roundedpicture[width = \wii]{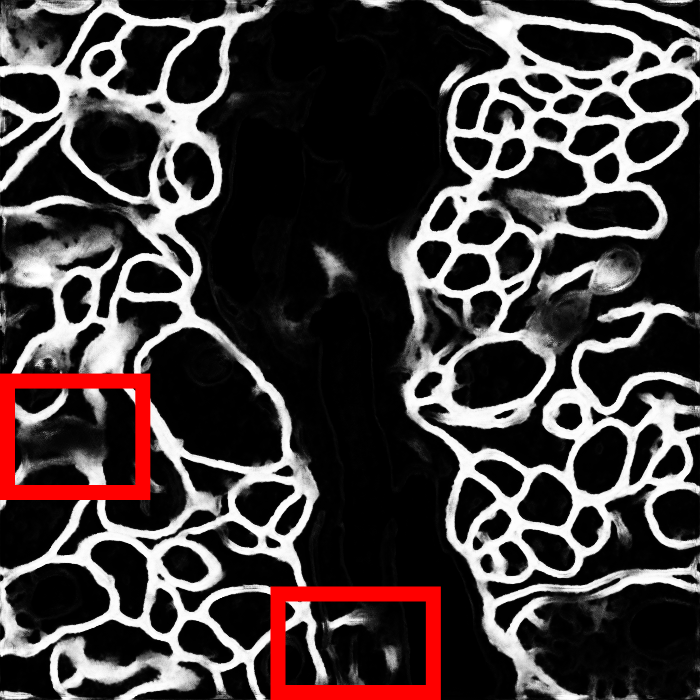}&\hspace{-3.5mm}
\roundedpicture[width = \wii]{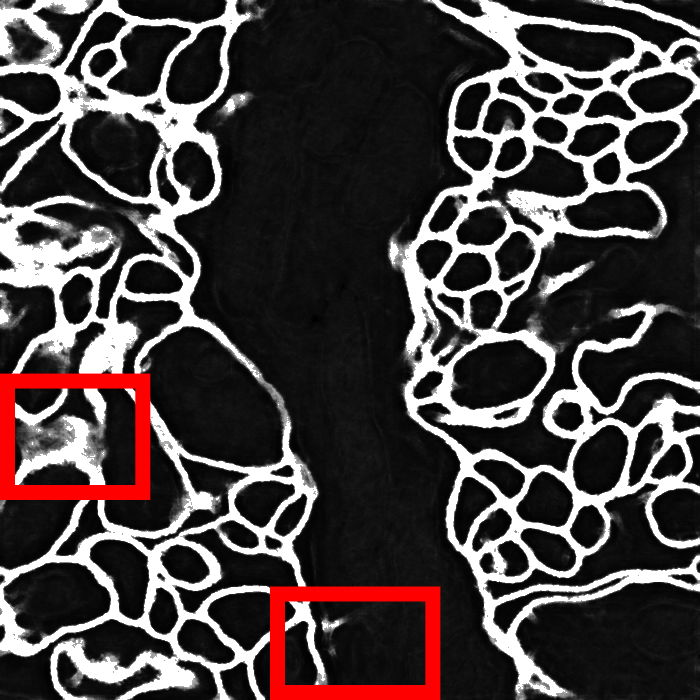}&\hspace{-3.5mm}
\roundedpicture[width = \wii]{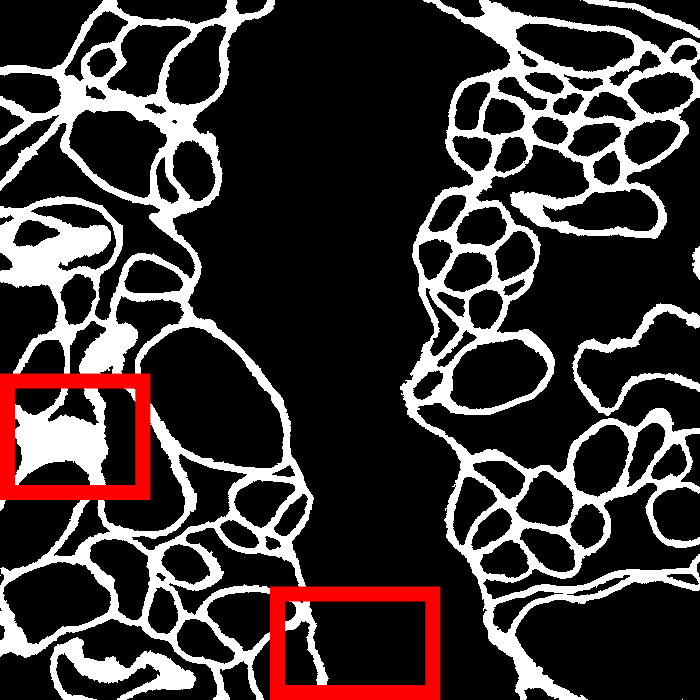} \\[-.5ex]
\roundedpicture[width = \wii]{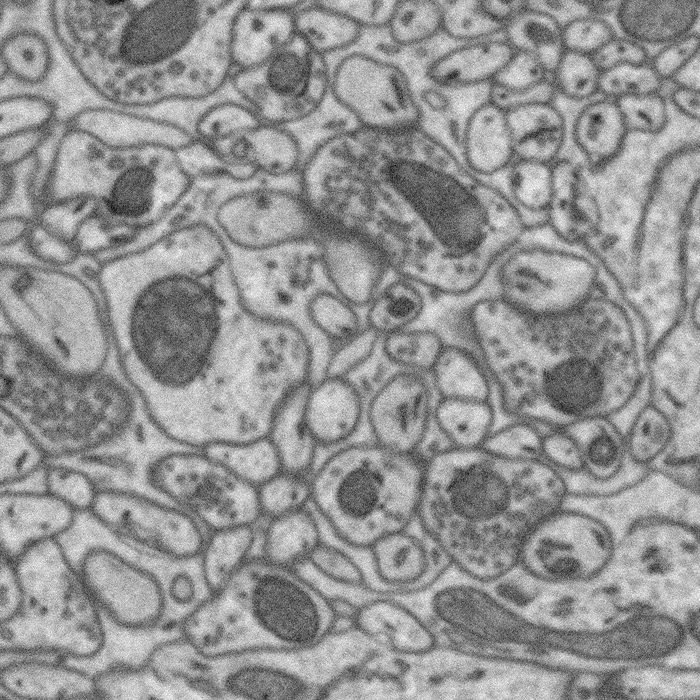}&\hspace{-3.5mm}
\roundedpicture[width = \wii]{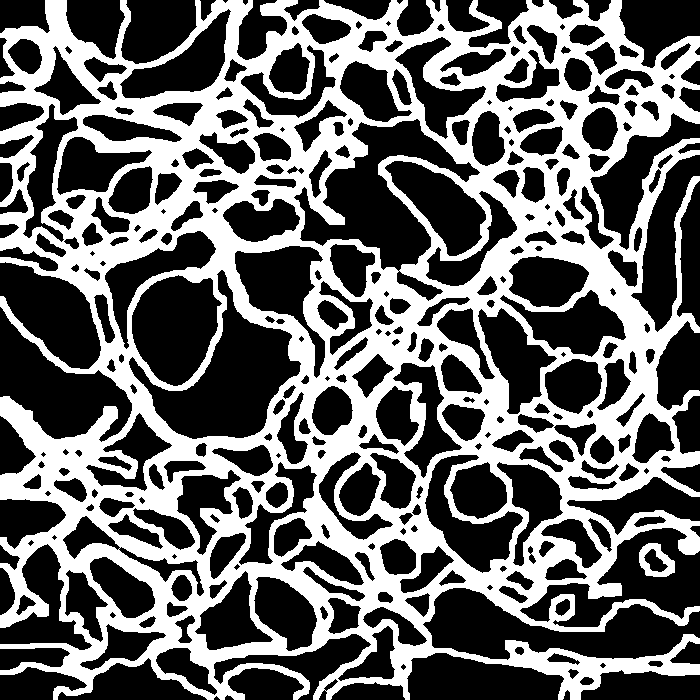} &\hspace{-3.5mm}
\roundedpicture[width = \wii]{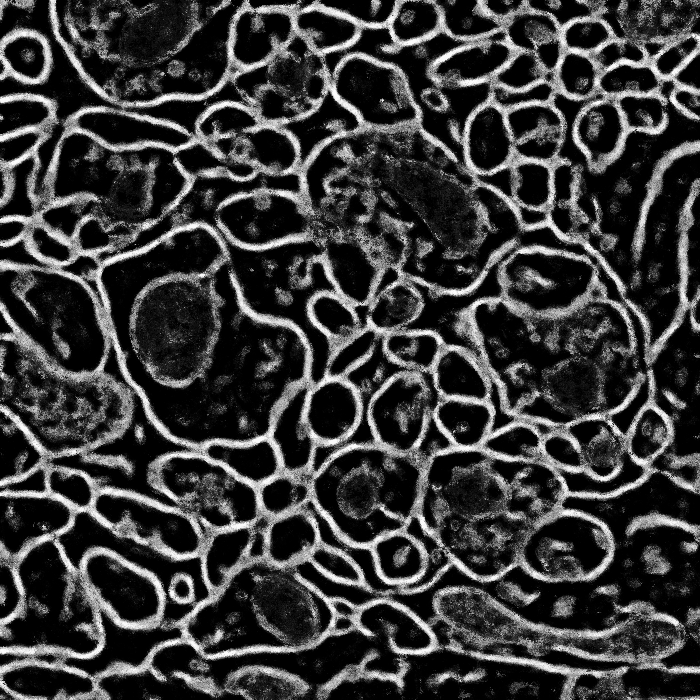}&\hspace{-3.5mm}
\roundedpicture[width = \wii]{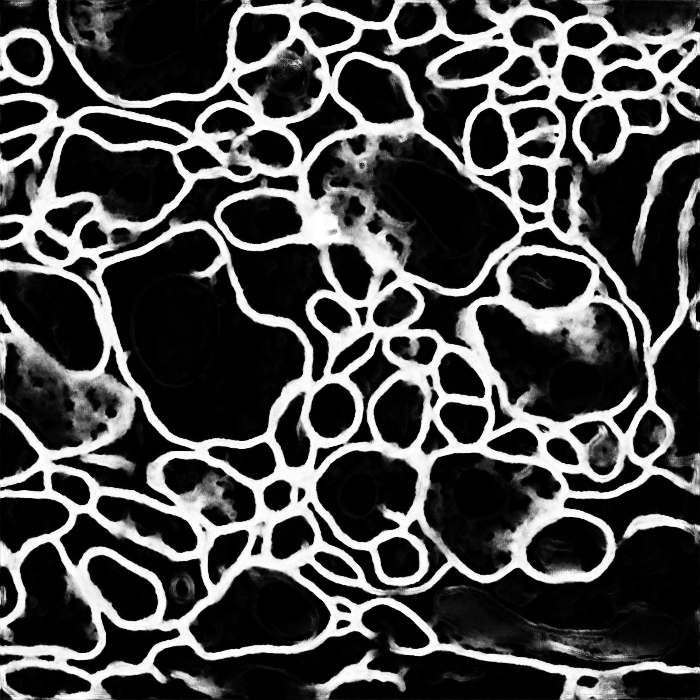}&\hspace{-3.5mm}
\roundedpicture[width = \wii]{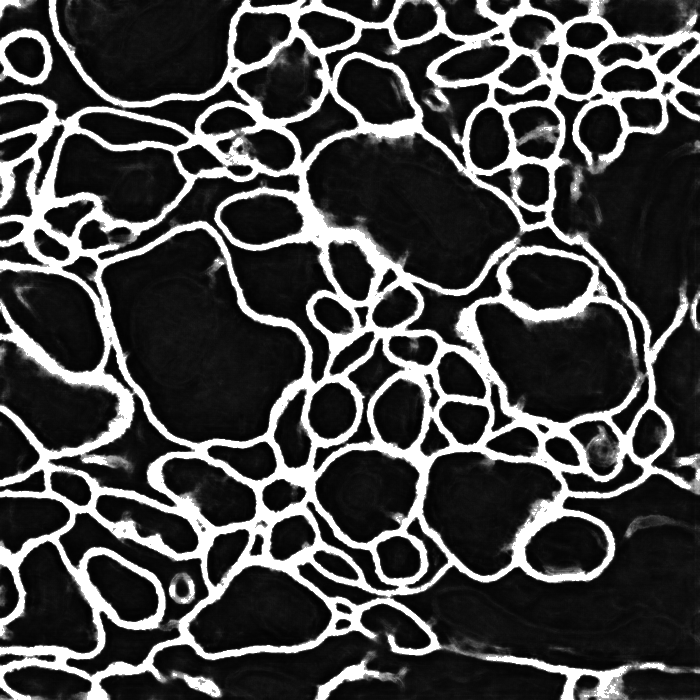}&\hspace{-3.5mm}
\roundedpicture[width = \wii]{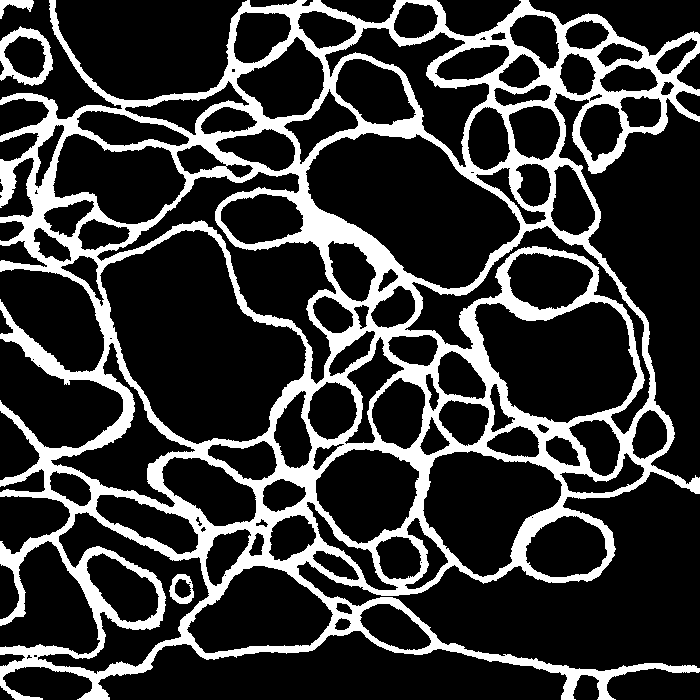} \\[-.5ex]
\roundedpicture[width = \wii]{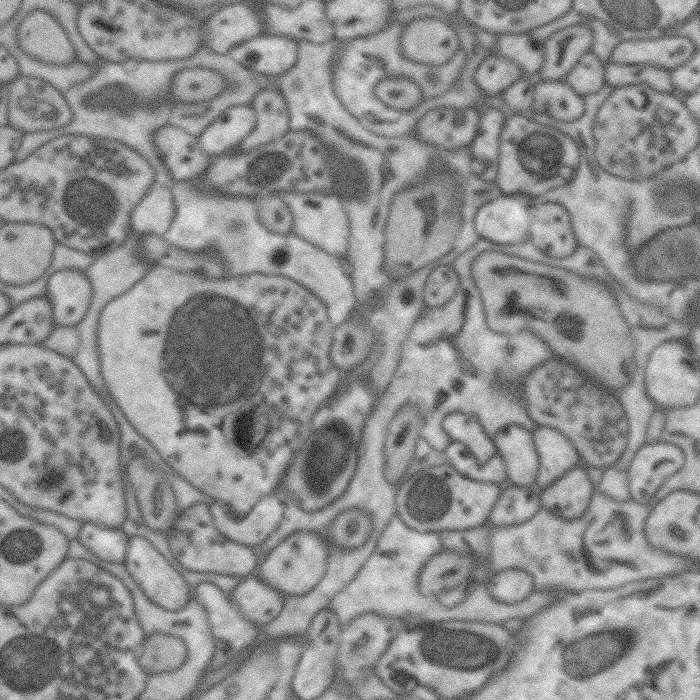}&\hspace{-3.5mm}
\roundedpicture[width = \wii]{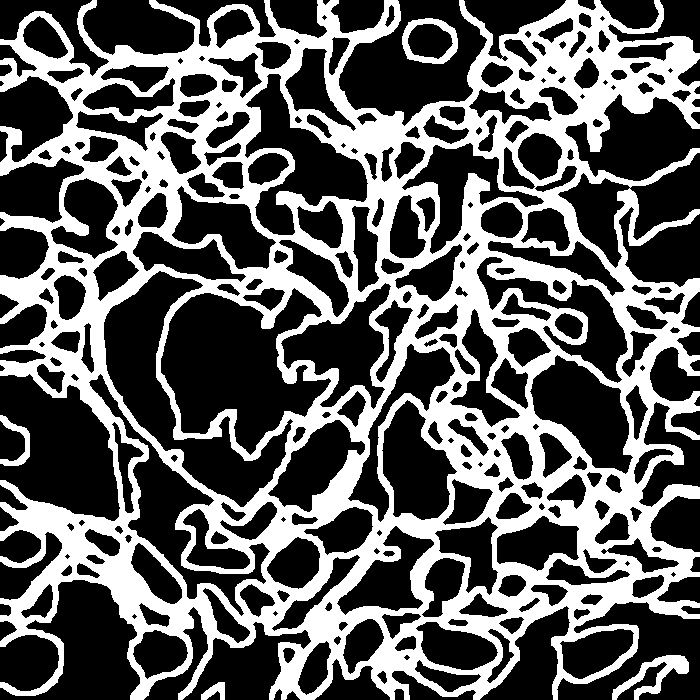} &\hspace{-3.5mm}
\roundedpicture[width = \wii]{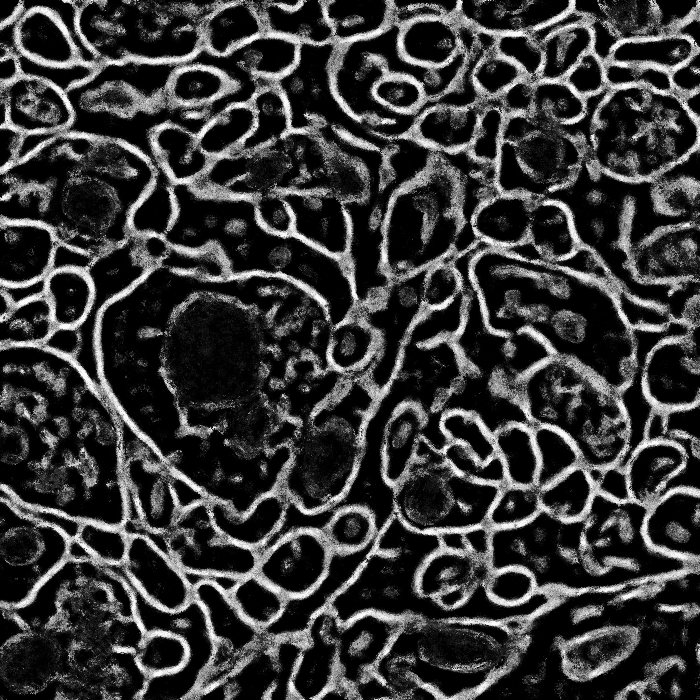}&\hspace{-3.5mm}
\roundedpicture[width = \wii]{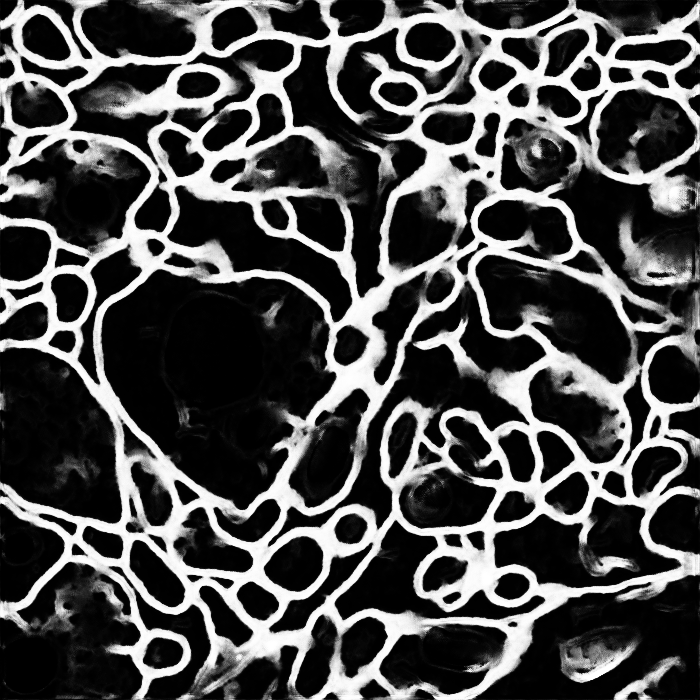}&\hspace{-3.5mm}
\roundedpicture[width = \wii]{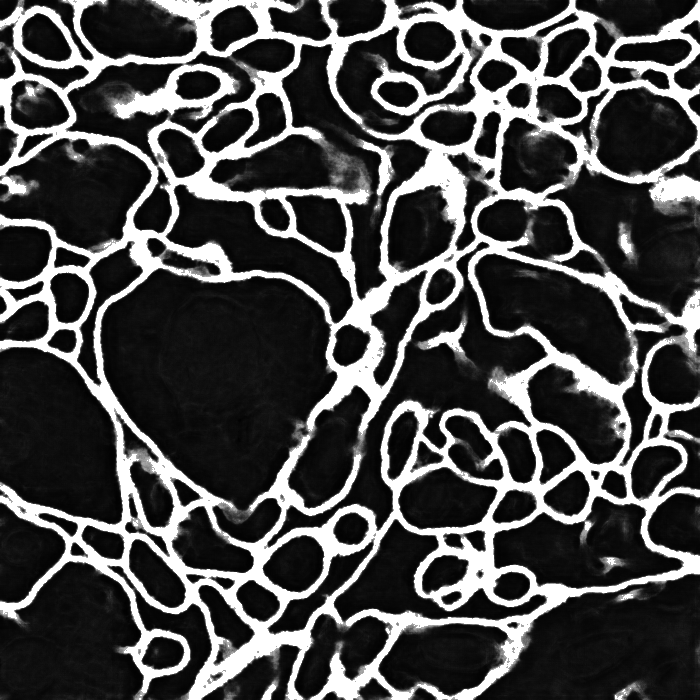}&\hspace{-3.5mm}
\roundedpicture[width = \wii]{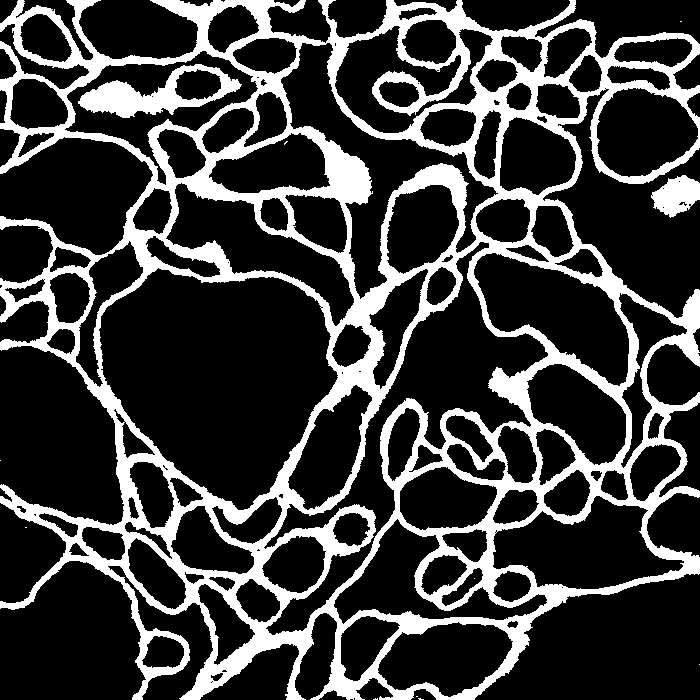} \\[-.5ex]
\roundedpicture[width = \wii]{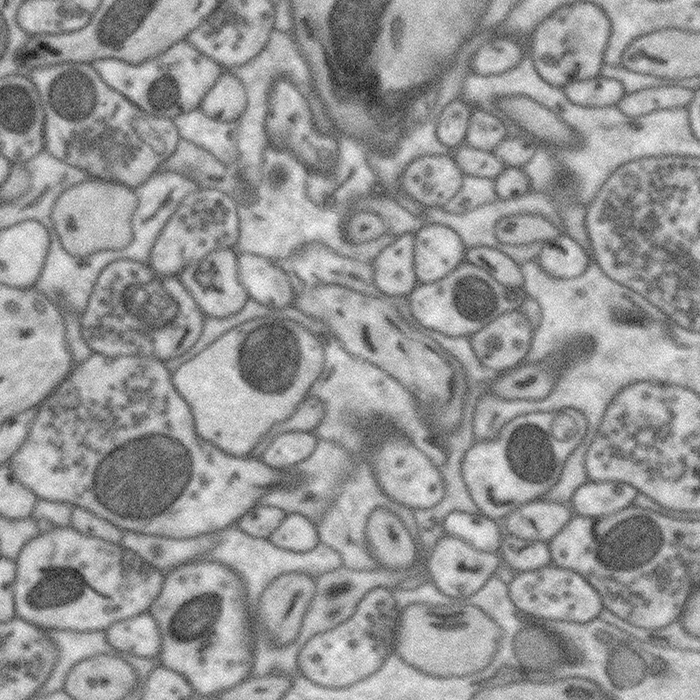}&\hspace{-3.5mm}
\roundedpicture[width = \wii]{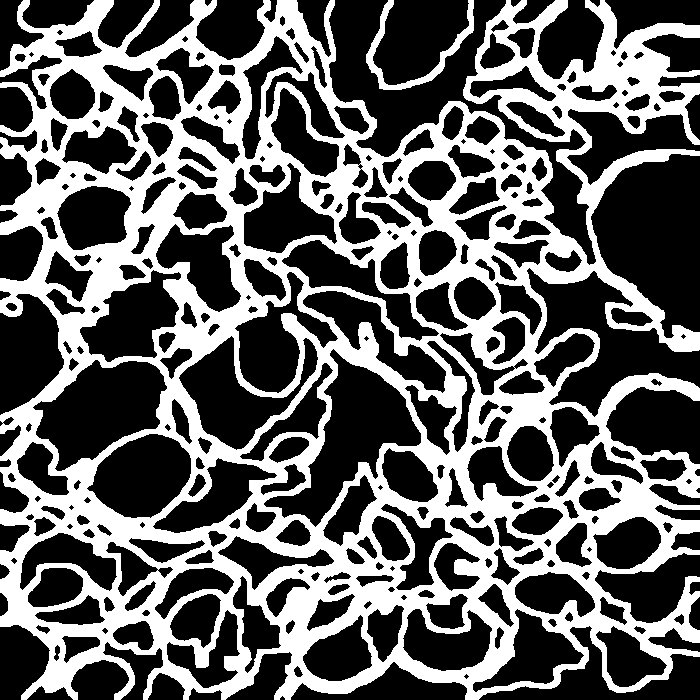} &\hspace{-3.5mm}
\roundedpicture[width = \wii]{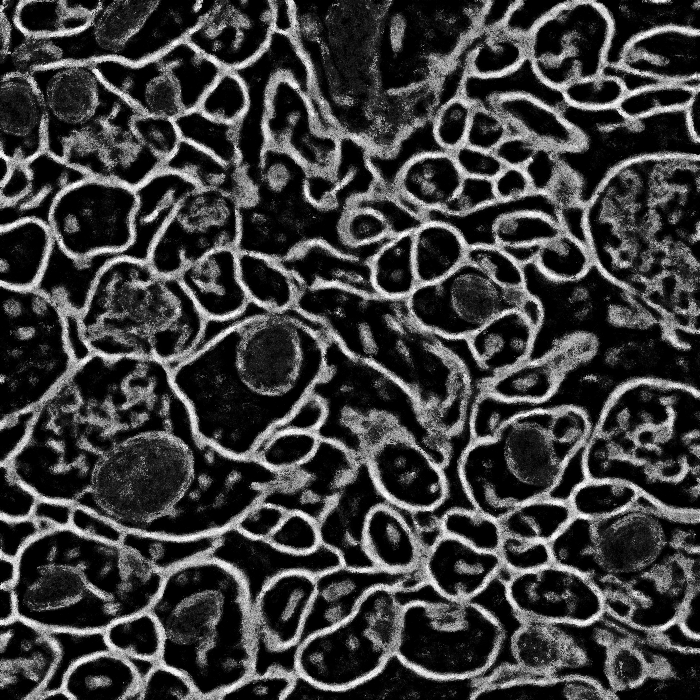}&\hspace{-3.5mm}
\roundedpicture[width = \wii]{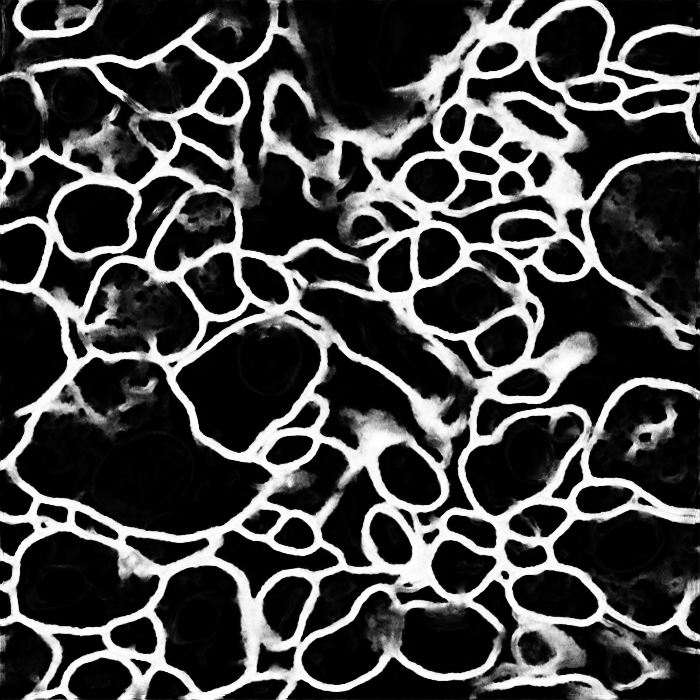}&\hspace{-3.5mm}
\roundedpicture[width = \wii]{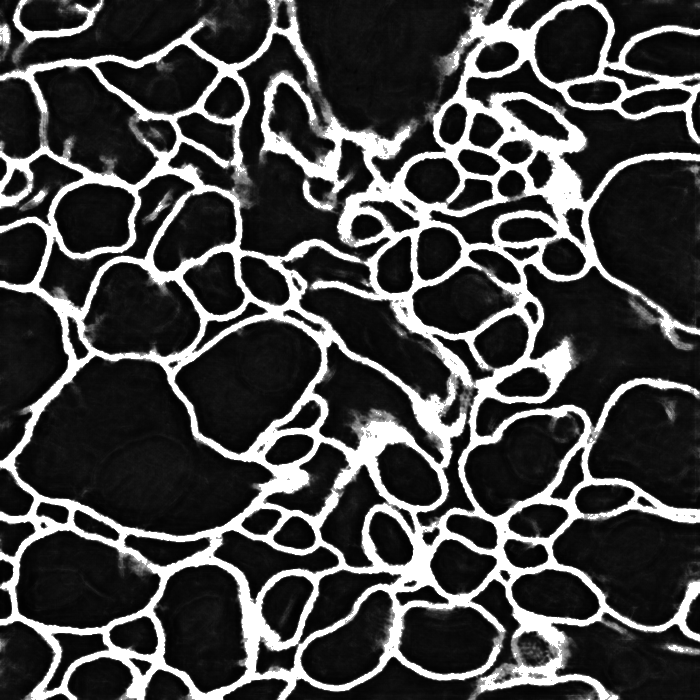}&\hspace{-3.5mm}
\roundedpicture[width = \wii]{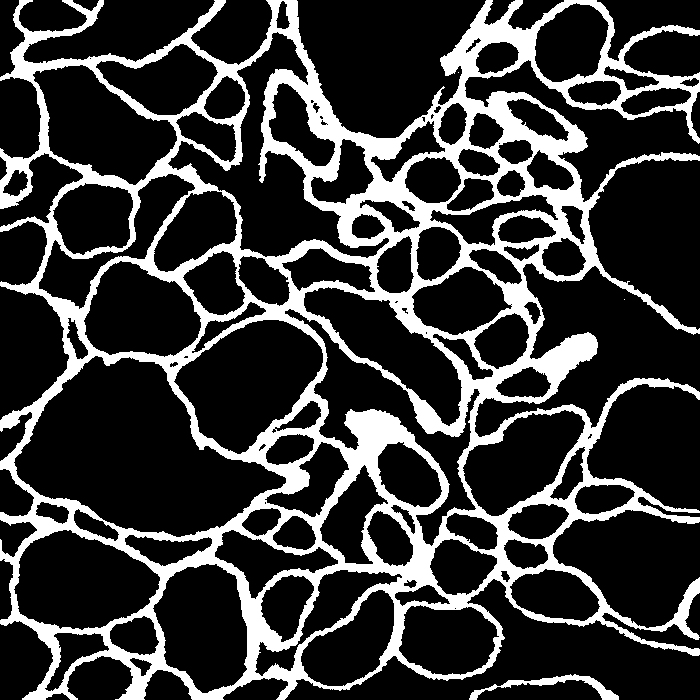}  \\[-.5ex]
(a)&(b)&(c)&(d)&(e)&(f)\\

\end{tabular}
\caption{Test results of the mouse neuropil dataset. (a) Input image, (b) gPb-OWT-UCM~\cite{Arbelaez2009}, (c) BEL~\cite{Dollar2006}, (d) MSANN~\cite{Seyedhosseini2011}, (e) CHM-LDNN, (f) ground truth images. The CHM is more successful in removing undesired parts and closing small gaps. Some of the improvements are marked with red rectangles.
For gPb-OWT-UCM method, the best threshold was picked and the edges
were dilated to the true membrane thickness.}
\label{Fig:mouseneuropil}

\end{figure*}

A few examples of the test images and corresponding membrane detection results using different methods are shown in Figure~\ref{Fig:mouseneuropil}. As shown in our results, the CHM outperforms MSANN in removing undesired parts from the background and closing some gaps.
\subsection{Drosophila VNC dataset}\label{Sec:VNC}
\begin{figure*}
\centering
\def \wii{.133\linewidth}
\begin{tabular}{ccccccc}

\roundedpicture[width = \wii]{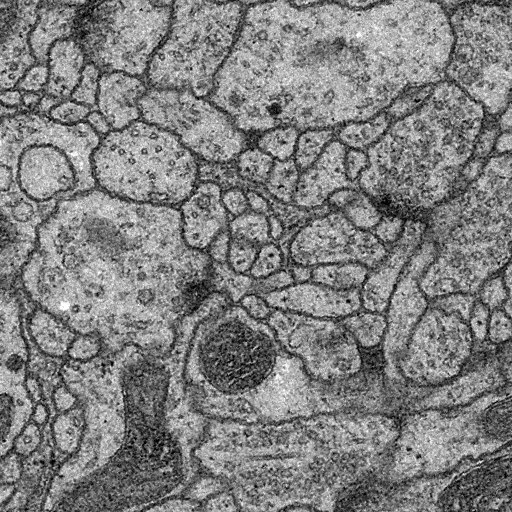}&\hspace{-3.5mm}
\roundedpicture[width = \wii]{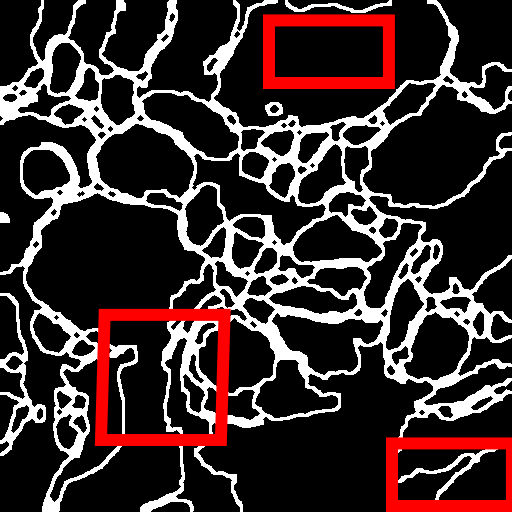} &\hspace{-3.5mm}
\roundedpicture[width = \wii]{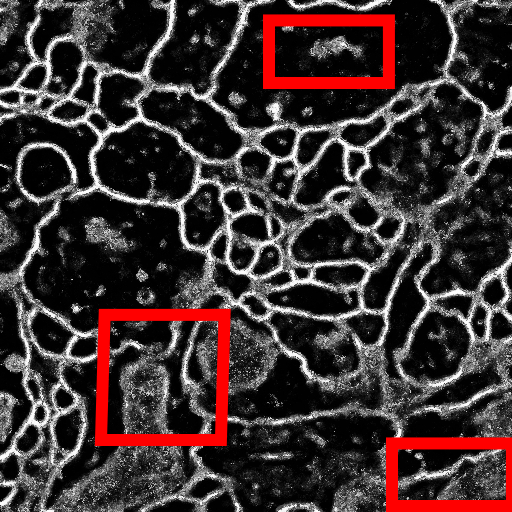}&\hspace{-3.5mm}
\roundedpicture[width = \wii]{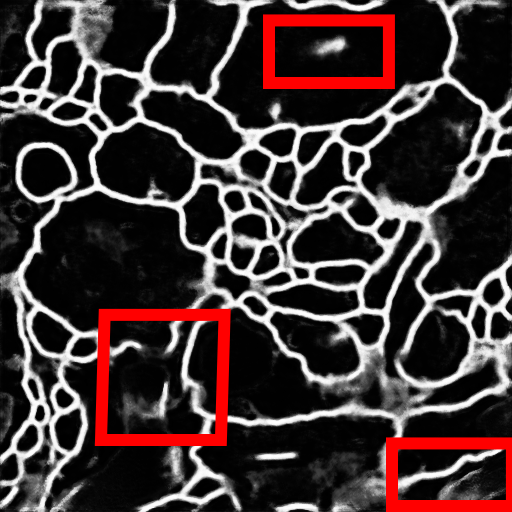}&\hspace{-3.5mm}
\roundedpicture[width = \wii]{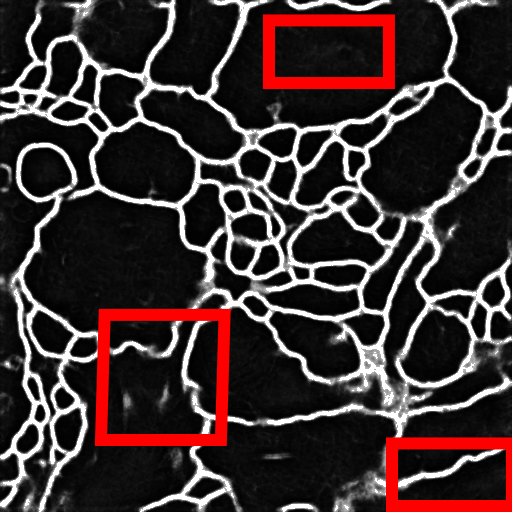}&\hspace{-3.5mm}
\roundedpicture[width = \wii]{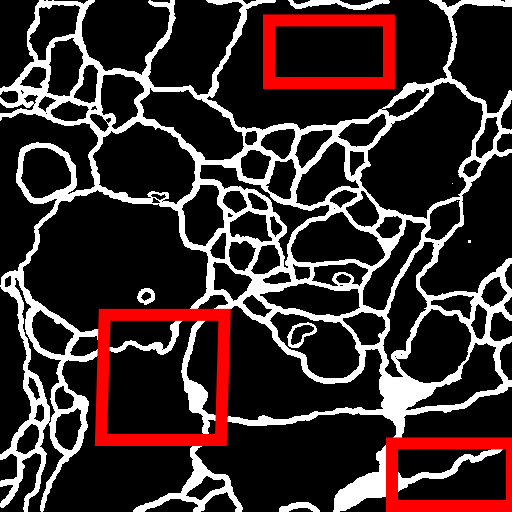}\\[-.5ex]
\roundedpicture[width = \wii]{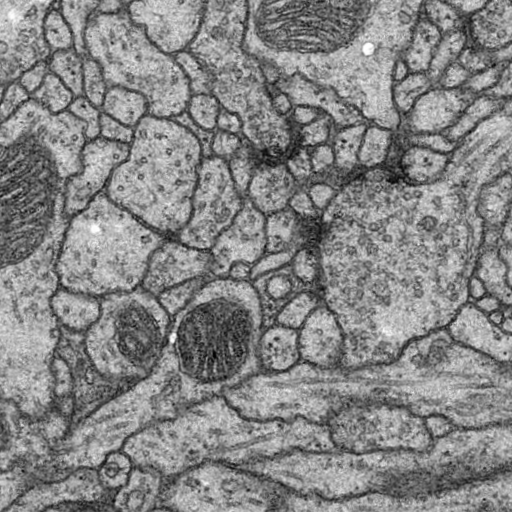}&\hspace{-3.5mm}
\roundedpicture[width = \wii]{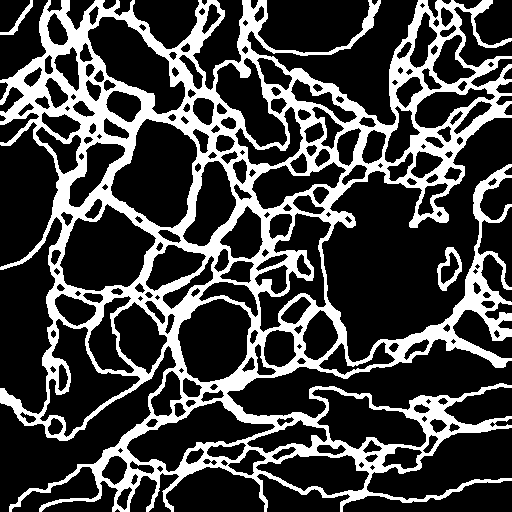} &\hspace{-3.5mm}
\roundedpicture[width = \wii]{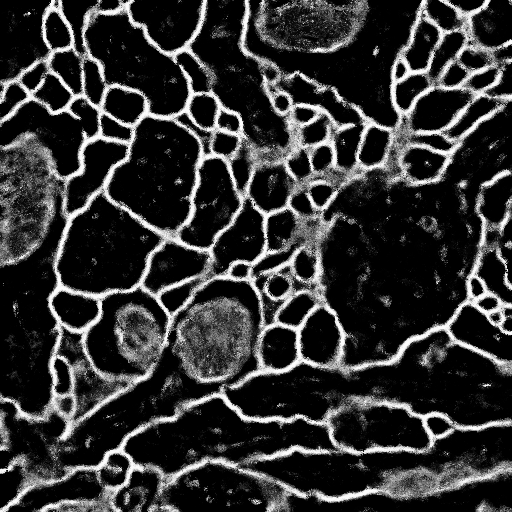}&\hspace{-3.5mm}
\roundedpicture[width = \wii]{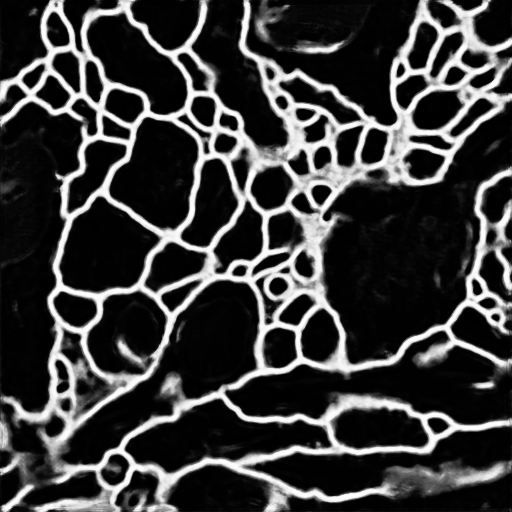}&\hspace{-3.5mm}
\roundedpicture[width = \wii]{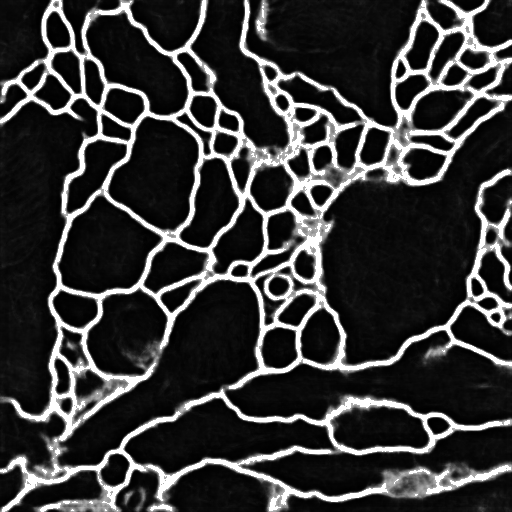}&\hspace{-3.5mm}
\roundedpicture[width = \wii]{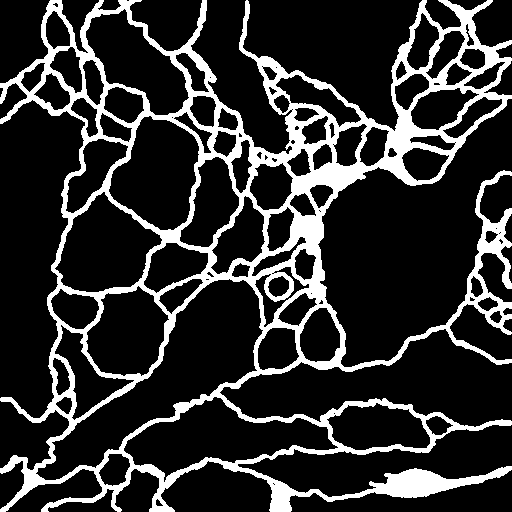}\\[-.5ex]
\roundedpicture[width = \wii]{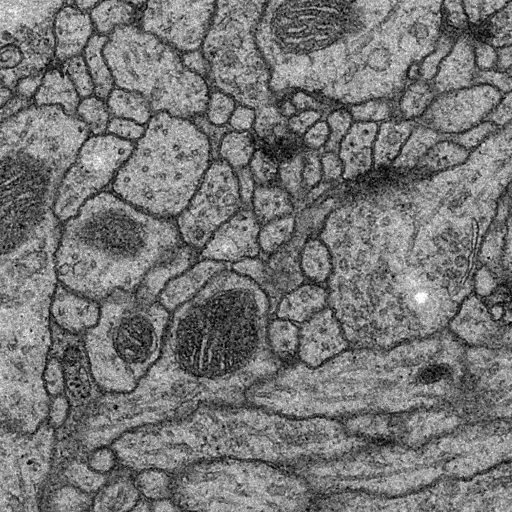}&\hspace{-3.5mm}
\roundedpicture[width = \wii]{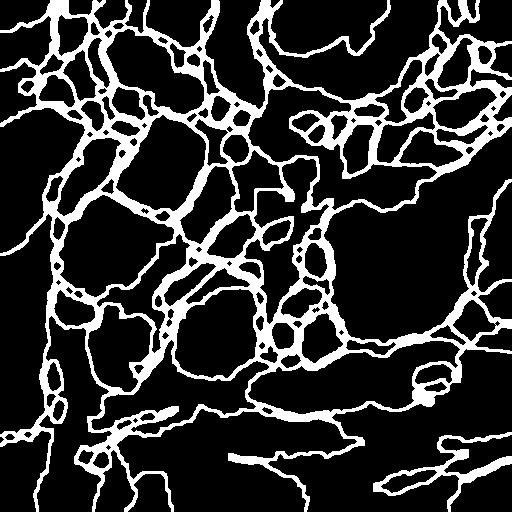} &\hspace{-3.5mm}
\roundedpicture[width = \wii]{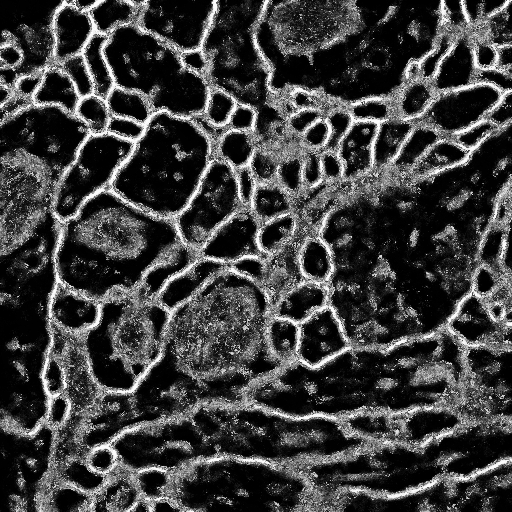}&\hspace{-3.5mm}
\roundedpicture[width = \wii]{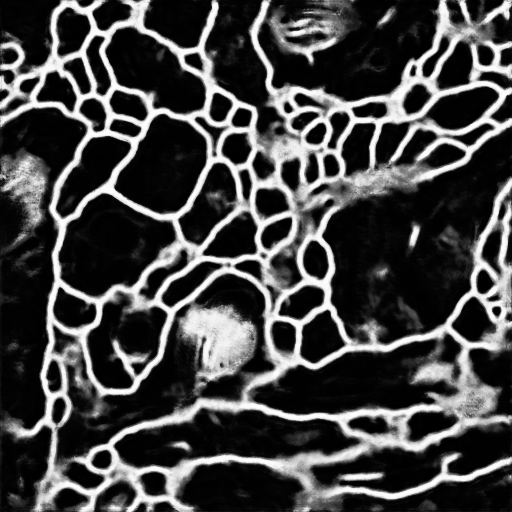}&\hspace{-3.5mm}
\roundedpicture[width = \wii]{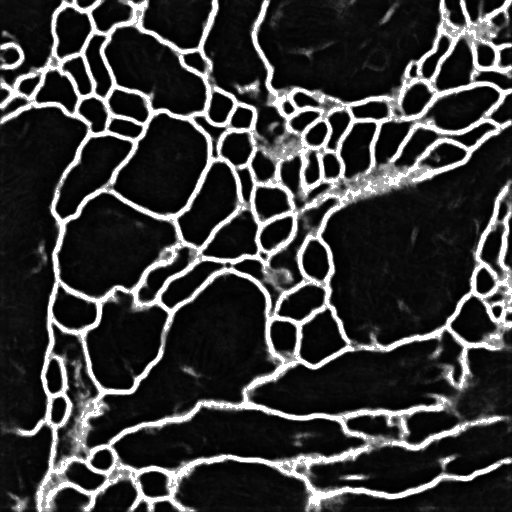}&\hspace{-3.5mm}
\roundedpicture[width = \wii]{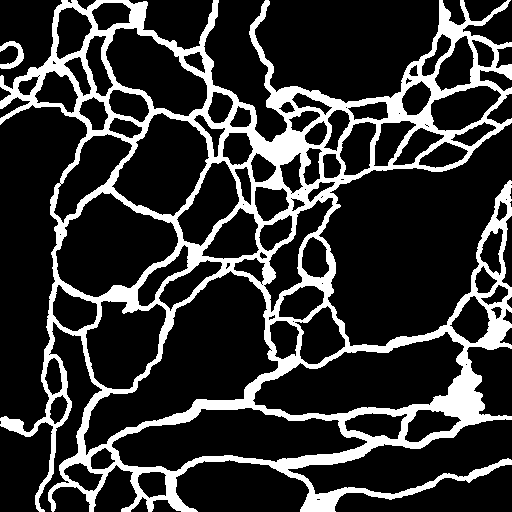}\\[-.5ex]
\roundedpicture[width = \wii]{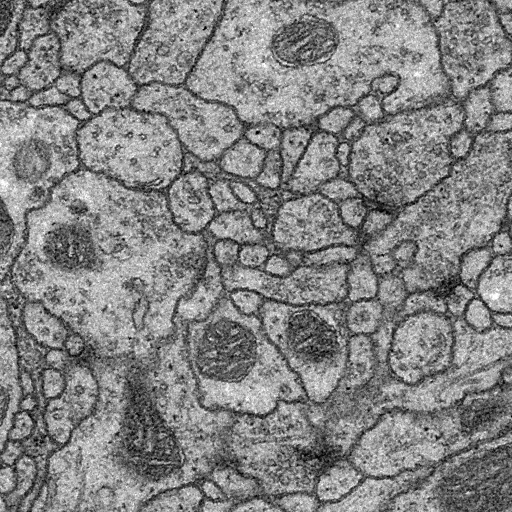}&\hspace{-3.5mm}
\roundedpicture[width = \wii]{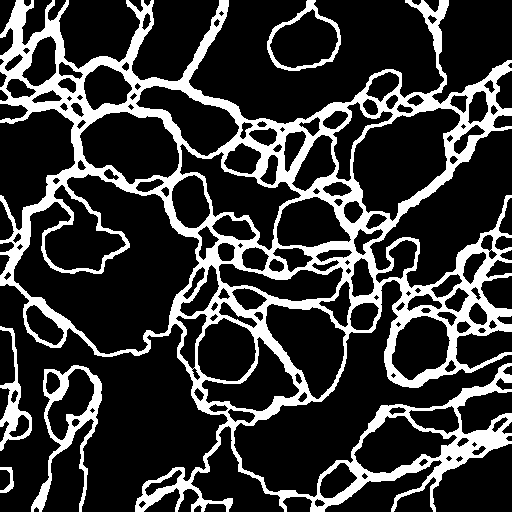} &\hspace{-3.5mm}
\roundedpicture[width = \wii]{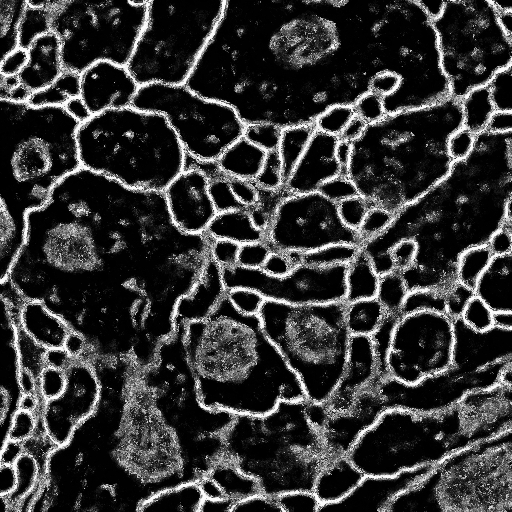}&\hspace{-3.5mm}
\roundedpicture[width = \wii]{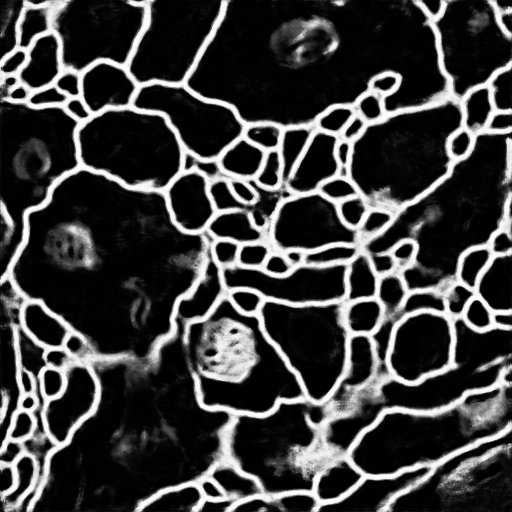}&\hspace{-3.5mm}
\roundedpicture[width = \wii]{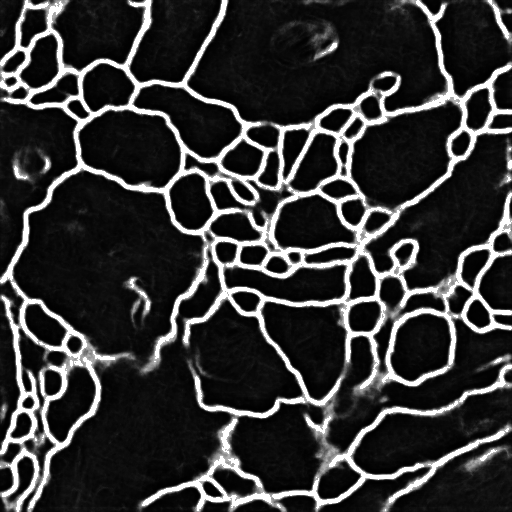}&\hspace{-3.5mm}
\roundedpicture[width = \wii]{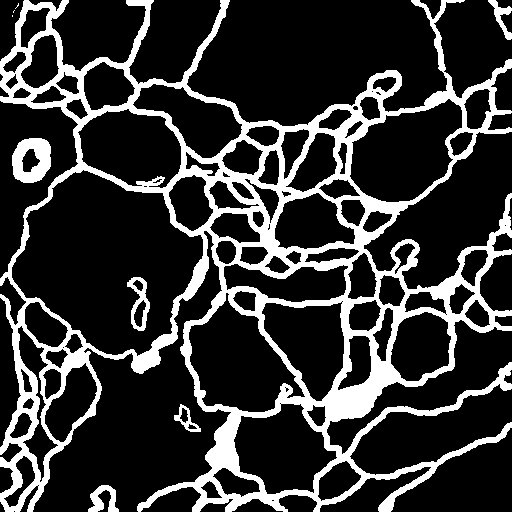}\\[-.5ex]
(a)&(b)&(c)&(d)&(e)&(f)\\

\end{tabular}
\caption{Test results of the Drosophila VNC dataset (second row). (a) Input image, (b) gPb-OWT-UCM~\cite{Arbelaez2009}, (c) BEL~\cite{Dollar2006}, (d) MSANN~\cite{Seyedhosseini2011}, (e) CHM, (f) ground truth images. The CHM is more successful in removing undesired parts and closing small gaps. Some of the improvements are marked with red rectangles.
For gPb-OWT-UCM method, the best threshold was picked and the edges
were dilated to the true membrane thickness.}
\label{Fig:Cardona}

\end{figure*}

This dataset contains $30$ images from Drosophila first instar larva ventral nerve cord (VNC)~\cite{Cardona2010,Cardona2012} acquired using serial-section transmission electron microscopy (ssTEM~\cite{Anderson2009,Chklovskii2010}).
Each image is $512$ by $512$ pixels and the resolution is  $4 \times 4 \times 50$ nm/pixel.
The membranes are marked by a human expert in each image.
We used $15$ images for training and $15$ images for testing.
The testing performance for different methods are reported in Table~\ref{Tab:NCMIR}.
It can be seen that the CHM outperforms the other methods in terms of pixel error.
A few test samples and membrane detection results for different methods are shown in Figure~\ref{Fig:Cardona}.

The same dataset was used as the training set for the ISBI 2012 EM challenge~\cite{ISBI2012}.
The participants were asked to submit the results on a different test set (the same size as the training set) to the challenge server.
We trained the same model on the whole $30$ images and submitted the results for the testing volume to the challenge server~\cite{ISBI2012}. The pixel error ($1-$F-value) of different methods are reported in Table~\ref{Tab:ISBIch}.
CHM achieved pixel error of $0.063$ which is better than the human error, \textit{i.e.,} how much a second human labeling differed from the first one.
It also outperformed the convolutional networks proposed in~\cite{Ciresan2012Deep} and~\cite{Jain2007}.
It is noteworthy that CHM is significantly faster than deep neural networks (DNN)~\cite{Ciresan2012Deep} at training. While DNN needs $85$ hours on GPU for training, CHM only needs $30$ hours on CPU.
\begin{table}
\caption{Pixel error ($1-$F-value) and training time (hours) of different methods on ISBI challenge~\cite{ISBI2012} test set.  Numbers are available on the challenge leader board.}
\centering
\begin{tabular}{ccc} \toprule
Method & $1-$F-value & Training Time \\ \midrule
Laptev \textit{et al.}~\cite{Laptev2012}&$0.067$& $?$\\ \midrule
Convolutional Networks~\cite{Jain2007}&$0.067$&$?$\\ \midrule
Human&$0.066$ &$-$\\ \midrule
Deep Neural Networks~\cite{Ciresan2012Deep}&$0.065$& $85 (GPU)$\\ \midrule
CHM&$\bf{0.063}$ &$30 (CPU)$\\ \bottomrule
\end{tabular}
\label{Tab:ISBIch}
\end{table}

\section{Conclusion and future work}

We develop a discriminative learning scheme for scene labeling, called CHM, which takes advantage of contextual information at multiple resolutions in a hierarchy.
The main advantage of CHM is its ability to optimize a posterior probability at multiple resolutions.
To our knowledge, this is the first time that a posterior at multiple resolutions is optimized for scene labeling.
CHM performs this optimization efficiently in a greedy manner.
To achieve this goal, CHM trains several classifiers at multiple resolutions and leverages the obtained results for learning a classifier at the original resolution.
We applied our model to several challenging datasets for scene labeling, edge detection, and biomedical image segmentation.
Results indicate that CHM achieves state-of-the-art performance on all of these applications.

An important characteristic of CHM is that it is only based on patch information and does not make use of any exemplars or shape models.
This enables CHM to serve as a general labeling method with high accuracy.
The other advantage of CHM is its simple training.
Even though our model needs to learn hundreds of parameters, the training remains tractable since classifiers are trained one at a time separately.

We conclude by discussing a possible extension of the CHM.
Even though CHM is able to model global contextual information within a scene, it can be prone to error due to absence of any global constrains.
Therefore, CHM can be used as a first step in a scene labeling pipeline.
Postprocessing such as CRF proposed in~\cite{Farabet2013} can be used to enforce label consistency and global constraints

\ifCLASSOPTIONcompsoc
  \section*{Acknowledgments}
\else
  \section*{Acknowledgment}
\fi

This work was supported by NIH 1R01NS075314-01 (TT,MHE) and NSF IIS-1149299(TT). We thank the ``National Center for Microscopy Imaging Research'' and the ``Cardona Lab at HHMI Janelia Farm'' for providing the mouse neuropil and Drosophila VNC datasets. We also thank Piotr Dollar for providing edge detection results of SE~\cite{Dollar2013} method for NYU depth dataset.

\ifCLASSOPTIONcaptionsoff
  \newpage
\fi



%
\bibliographystyle{IEEEtran}
\bibliography{egbib}

\end{document}